\documentclass[10pt,twocolumn]{article}

\usepackage[margin=0.75in]{geometry}
\usepackage{booktabs}
\usepackage{graphicx}
\usepackage{amsmath}
\usepackage{xcolor}
\usepackage{multirow}
\usepackage{array}
\usepackage{longtable}
\usepackage{caption}
\usepackage{enumitem}
\usepackage{microtype}
\usepackage{adjustbox}
\usepackage[most]{tcolorbox}
\usepackage{authblk}

\usepackage[hidelinks]{hyperref}

\usepackage{dblfloatfix}

\usepackage[style=authoryear,natbib=true,backend=biber]{biblatex}
\newcommand{\CorpusRecords}{1185}
\newcommand{\CorpusEntries}{5230}
\newcommand{\CorpusGoldMean}{1.96}
\newcommand{\CorpusHnMean}{5.08}

\newcommand{\NormIndustryPharmaceuticals}{499}
\newcommand{\NormIndustryAutomotive}{248}
\newcommand{\NormIndustryOilgas}{438}

\newcommand{\TaxTotal}{6021}

\newcommand{\TaxContamRecords}{305}
\newcommand{\TaxContamHN}{442}
\newcommand{\BMRecallOne}{8.3}
\newcommand{\BMRecallFive}{24.5}
\newcommand{\BMRecallTen}{32.1}
\newcommand{\BMRecallTwenty}{40.1}
\newcommand{\BMMRR}{23.8}
\newcommand{\BMNDCG}{21.8}
\newcommand{\BMInRecMRR}{78.2}
\newcommand{\BMInRecNDCGFive}{75.4}

\newcommand{\TfidfRecallOne}{7.7}
\newcommand{\TfidfRecallFive}{23.7}
\newcommand{\TfidfRecallTen}{32.8}
\newcommand{\TfidfRecallTwenty}{41.9}
\newcommand{\TfidfMRR}{23.9}
\newcommand{\TfidfNDCG}{21.9}
\newcommand{\TfidfInRecMRR}{80.4}
\newcommand{\TfidfInRecNDCGFive}{77.9}

\newcommand{\DenseRecallOne}{6.5}
\newcommand{\DenseRecallFive}{21.5}
\newcommand{\DenseRecallTen}{31.8}
\newcommand{\DenseRecallTwenty}{42.0}
\newcommand{\DenseMRR}{22.2}
\newcommand{\DenseNDCG}{20.4}
\newcommand{\DenseInRecMRR}{78.6}
\newcommand{\DenseInRecNDCGFive}{74.7}

\newcommand{\CeDenseRecallOne}{9.6}
\newcommand{\CeDenseRecallFive}{27.5}
\newcommand{\CeDenseRecallTen}{35.2}
\newcommand{\CeDenseRecallTwenty}{42.0}
\newcommand{\CeDenseMRR}{26.9}
\newcommand{\CeDenseNDCG}{24.6}

\newcommand{\CeInrecInRecMRR}{79.8}
\newcommand{\CeInrecInRecNDCGFive}{76.6}

\newcommand{\AblDoctypeTenKBmTwofiveRecallTen}{33.9}

\newcommand{\AblDoctypeTenKDenseRecallTen}{33.7}

\newcommand{\AblDoctypeTenKCeDenseTopKRecallTen}{36.9}

\newcommand{\AblDoctypeTenQBmTwofiveRecallTen}{18.0}

\newcommand{\AblDoctypeTenQDenseRecallTen}{17.3}

\newcommand{\AblDoctypeTenQCeDenseTopKRecallTen}{21.8}

\newcommand{\AblReasoningtypeQualitativeBmTwofiveRecallTen}{34.7}

\newcommand{\AblReasoningtypeQualitativeDenseRecallTen}{35.2}

\newcommand{\AblReasoningtypeQualitativeCeDenseTopKRecallTen}{38.2}

\newcommand{\AblReasoningtypeQuantitativeBmTwofiveRecallTen}{26.5}

\newcommand{\AblReasoningtypeQuantitativeDenseRecallTen}{24.4}

\newcommand{\AblReasoningtypeQuantitativeCeDenseTopKRecallTen}{28.8}

\newcommand{\AblPassagetypeSinglepassageBmTwofiveRecallTen}{44.7}

\newcommand{\AblPassagetypeSinglepassageDenseRecallTen}{42.3}

\newcommand{\AblPassagetypeSinglepassageCeDenseTopKRecallTen}{46.9}

\newcommand{\AblPassagetypeTwopassageBmTwofiveRecallTen}{28.0}

\newcommand{\AblPassagetypeTwopassageDenseRecallTen}{28.8}

\newcommand{\AblPassagetypeTwopassageCeDenseTopKRecallTen}{31.7}

\newcommand{\AblPassagetypeMultipassageBmTwofiveRecallTen}{21.5}

\newcommand{\AblPassagetypeMultipassageDenseRecallTen}{22.2}

\newcommand{\AblPassagetypeMultipassageCeDenseTopKRecallTen}{24.8}

\newcommand{\AblDifficultyEasyBmTwofiveRecallTen}{37.6}

\newcommand{\AblDifficultyEasyDenseRecallTen}{33.7}

\newcommand{\AblDifficultyEasyCeDenseTopKRecallTen}{39.0}

\newcommand{\AblDifficultyMediumBmTwofiveRecallTen}{32.4}

\newcommand{\AblDifficultyMediumDenseRecallTen}{32.5}

\newcommand{\AblDifficultyMediumCeDenseTopKRecallTen}{35.3}

\newcommand{\AblDifficultyHardBmTwofiveRecallTen}{25.9}

\newcommand{\AblDifficultyHardDenseRecallTen}{28.6}

\newcommand{\AblDifficultyHardCeDenseTopKRecallTen}{31.0}

\newcommand{\QRRawBmTwofiveRecallTen}{29.9}

\newcommand{\QRWithrewritesBmTwofiveRecallTen}{38.7}

\newcommand{\QRRawTfidfRecallTen}{32.3}

\newcommand{\QRWithrewritesTfidfRecallTen}{41.1}

\newcommand{\QRRawDenseRecallTen}{30.7}

\newcommand{\QRWithrewritesDenseRecallTen}{34.1}

\newcommand{\MetaFilterBmRecallTen}{55.0}
\newcommand{\MetaFilterBmMRR}{41.9}

\newcommand{\MetaFilterHnElimPct}{92.9}
\newcommand{\MetaFilterMeanCands}{173}
\newcommand{\MetaFilterGoldLossSome}{33}
\newcommand{\MetaFilterGoldLossAll}{2}
\newcommand{\SplitRandomTrain}{948}
\newcommand{\SplitRandomDev}{118}
\newcommand{\SplitRandomTest}{119}
\newcommand{\SplitTickerTrain}{910}
\newcommand{\SplitTickerDev}{140}
\newcommand{\SplitTickerTest}{135}
\newcommand{\SplitYearTest}{296}
\newcommand{\SplitDoctypeTest}{137}
\newcommand{\QRCount}{819}
\newcommand{\QRCoveragePct}{69}
\newcommand{\FinBgeRecallOne}{5.2}
\newcommand{\FinBgeRecallFive}{16.5}
\newcommand{\FinBgeRecallTen}{22.4}
\newcommand{\FinBgeRecallTwenty}{29.8}
\newcommand{\FinBgeMRR}{16.5}
\newcommand{\FinBgeNDCG}{14.8}
\newcommand{\FinBgeInRecMRR}{72.3}
\newcommand{\FinBgeInRecNDCGFive}{70.2}

\newcommand{\BgeLargeRecallOne}{8.8}
\newcommand{\BgeLargeRecallFive}{25.9}
\newcommand{\BgeLargeRecallTen}{35.6}
\newcommand{\BgeLargeRecallTwenty}{44.6}
\newcommand{\BgeLargeMRR}{25.9}
\newcommand{\BgeLargeNDCG}{23.8}
\newcommand{\BgeLargeInRecMRR}{79.1}
\newcommand{\BgeLargeInRecNDCGFive}{76.5}

\newcommand{\EFiveMistralRecallOne}{10.1}
\newcommand{\EFiveMistralRecallFive}{33.1}
\newcommand{\EFiveMistralRecallTen}{44.8}
\newcommand{\EFiveMistralRecallTwenty}{55.0}
\newcommand{\EFiveMistralMRR}{31.0}
\newcommand{\EFiveMistralNDCG}{29.7}
\newcommand{\EFiveMistralInRecMRR}{83.6}
\newcommand{\EFiveMistralInRecNDCGFive}{80.5}

\title{FinRank: An Evidence-Grounded Benchmark for Financial Question Answering and Retrieval over SEC Filings}

\author[1]{Sasan Mansouri\thanks{Corresponding author: \texttt{s.mansouri@rug.nl}}}
\author[2]{Daniel Saad}
\author[2]{Mark Wahrenburg}
\author[2,3]{Manu Weissel}
\author[4]{Fabian Woebbeking}
\affil[1]{University of Groningen, Groningen, Netherlands}
\affil[2]{Goethe University Frankfurt, Frankfurt am Main, Germany}
\affil[3]{DataNXT GmbH, Frankfurt am Main, Germany}
\affil[4]{Halle Institute for Economic Research (IWH) and Martin Luther University Halle-Wittenberg, Halle (Saale), Germany}

\date{}

\begin{document}

\maketitle

\begin{abstract}
Financial question answering is typically evaluated by answer correctness, yet in SEC filings a plausible and even numerically correct answer can be grounded in the wrong evidence. Similar facts and disclosures recur across sections of a filing, across reporting periods of the same firm, and across comparable firms. \textbf{FinRank} targets this provenance-sensitive retrieval problem by requiring systems to identify evidence for the intended entity, reporting period, and disclosure context. The benchmark contains \CorpusRecords{} manually authored question--answer records over the 10-K and 10-Q filings of 22 companies. Each record includes a reference answer, gold supporting passages, and hand-curated \emph{hard negatives} drawn from confusable passages within filings, across reporting periods, and across comparable firms. FinRank evaluates passage retrieval, reranking, and hard-negative discrimination as separately measured tasks. Baseline results demonstrate the difficulty of this setting: among the evaluated systems, even a 7B instruction-tuned embedder reaches only 44.8\% Recall@10 on the pooled evidence corpus; sub-billion-parameter encoders gain at most 3.5 points over BM25, a finance-adapted embedder trails BM25 by 9.7 points, and pairwise accuracy falls by 13.0--20.5 percentage points when random negatives are replaced with the curated hard negatives. FinRank provides an evidence-first benchmark for developing financial question answering systems that are not only accurate but also grounded in the correct disclosure.
\end{abstract}

\section{Introduction}
\label{sec:introduction}

Financial disclosures submitted to the U.S. Securities and Exchange Commission (SEC), such as Form 10-K and 10-Q filings, serve as the primary source of truth for corporate financial analysis, auditing, and regulatory compliance \citep{wu2023bloomberggpt,islam2023financebench}. While recent advances in retrieval-augmented generation (RAG) have enabled large language models (LLMs) to query extensive document corpora \citep{lewis2020rag,karpukhin2020dpr}, financial document question answering presents a fundamental challenge distinct from open-domain settings: regulatory disclosures are heavily templated. Standardized accounting conventions, risk-factor disclosures, and revenue-recognition notes read almost identically across competing firms and reporting periods. Consequently, the primary bottleneck in automated financial analysis is rarely answer composition, but \emph{evidence discrimination}—distinguishing the target firm's disclosure from plausible distractors, such as a competitor's near-identical boilerplate or a prior period's filing. For an analyst, auditor, or compliance reviewer, an ungrounded or misattributed answer carries limited practical utility and introduces severe operational risk; yet existing financial benchmarks \citep{chen2021finqa,zhu2021tatqa,reddy2024docfinqa} focus primarily on numerical calculation over provided snippets or end-to-end correctness, failing to isolate evidence retrieval and hard-negative suppression.

Beyond boilerplate templating, document length and structural complexity pose additional hurdles. A single 10-K filing frequently spans hundreds of pages of dense prose, financial tables, and footnotes, mixing accounting, legal, and forward-looking language that requires domain expertise to interpret. Furthermore, answers frequently depend on evidence spread across non-adjacent tables, footnotes, and narrative sections rather than a single paragraph. General-domain QA benchmarks \citep[e.g.,][]{rajpurkar2016squad,kwiatkowski2019nq}, by contrast, treat supporting evidence as a secondary annotation—inverting the requirements of financial domain QA, where an answer's validity depends on strict provenance to its underlying disclosure.

To address these challenges, we introduce \textbf{FinRank}, an evidence-grounded benchmark consisting of \CorpusRecords{} question--answer records manually authored over the 10-K and 10-Q filings of 22 companies across three sectors (pharmaceuticals, oil and gas, and automotive) covering filing years 2024--2025. Each record pairs a question and reference answer with gold supporting passages, filing and question-level metadata (topic, difficulty, reasoning type, evidence scope), and a hand-curated set of hard-negative passages drawn from comparable filings. In a realistic analyst workflow, a model must identify the passages that actually support an answer, distinguish them from plausible distractors, and produce a faithful response; FinRank evaluates the ranking and discrimination steps directly, over a curated passage pool. Its construction criteria (Section~\ref{sec:methodology}; Appendix~\ref{app:methodology}) jointly control topic coverage, qualitative versus quantitative reasoning, complexity, evidence scope, and sub-question decomposition, thereby ensuring that the benchmark evaluates financial reasoning and hard-negative discrimination rather than superficial lexical matching alone. FinRank is designed specifically for retrieval-grounded financial QA over selected SEC filings; its scope is defined by the included corporate disclosures, and it is not intended for generating automated investment advice, valuations, or trading signals.

In summary, our contributions are fourfold. First, we release FinRank\footnote{Dataset, evaluation harness, repair log, and hard-negative taxonomy: \url{https://github.com/datanxt/FinRank}} as, to our knowledge, the first financial QA benchmark to release a dedicated set of human-selected, semantically confusable hard negatives for each question, enabling direct comparison of discrimination against curated versus random distractors, complementing resources that focus on realistic queries \citep{choi2025finder} or open-book correctness \citep{islam2023financebench} without releasing gold hard negatives. Second, we provide rich, stratifiable annotations across four axes (document type, reasoning type, evidence scope, and difficulty) alongside gold sub-question decompositions (\texttt{query\_rewrite}) for \QRCoveragePct\% of records, enabling performance to be disaggregated along fine-grained dimensions obscured by aggregate metrics. Third, we establish reference retrieval and reranking baselines spanning sparse (TF-IDF, BM25), dense (bi-encoder), and cross-encoder architectures, demonstrating through hard-versus-random negative contrasts that curated distractors degrade model accuracy by 13.0--20.5 percentage points. Fourth, we support transparency, auditability, and extensibility through documented construction criteria, a deterministic label-normalization and repair pipeline with a machine-readable change log, per-passage identifiers and text hashes, a hard-negative taxonomy, and a target-versus-realized distributional audit.

\section{Related Work}
\label{sec:related-work}

\subsection{Financial QA and RAG Benchmarks}

Early financial QA benchmarks target numerical reasoning over a
\emph{provided} context. FinQA \citep{chen2021finqa} pairs questions with
executable reasoning programs over S\&P~500 earnings-report excerpts;
TAT-QA \citep{zhu2021tatqa} covers hybrid table--text reasoning;
ConvFinQA \citep{chen2022convfinqa} extends FinQA to multi-turn dialogue;
and MultiHiertt \citep{zhao2022multihiertt} and PACIFIC
\citep{deng2022pacific} push multi-table and proactive-conversational
reasoning. In all of these the supporting context is supplied, so
retrieval is not evaluated. A second line makes retrieval part of the
task: FinanceBench \citep{islam2023financebench} scores open-book answer
correctness over whole filings; DocFinQA \citep{reddy2024docfinqa}
lengthens FinQA to full-document context; FinTextQA
\citep{chen2024fintextqa} and T2-RAGBench \citep{strich2026t2ragbench}
benchmark end-to-end RAG over financial prose and tables; and FinDER
\citep{choi2025finder} contributes expert-written, abbreviation-heavy
queries with annotated evidence, scored with RAGAS-style metrics
\citep{es2023ragas}. The most recent entries extend the setting further
still, to agentic document- and chunk-ranking
\citep{choi2025finagentbench}, multimodal visual citation
\citep{zhao2025finragbenchv}, and analyst-workflow evaluation across
within-disclosure, cross-entity, and longitudinal pathways
\citep{zhu2026finrate}. FinDER itself originated as
the namesake task of the ICAIF'24 FinanceRAG Challenge
\citep{choi2024financerag}, which unified seven of these datasets into a
single NDCG@10 retrieval leaderboard. Beyond document-centric RAG, recent
work explores bypassing retrieval altogether by giving models direct
access to curated vendor data through the Model Context Protocol, which is
effective for quantitative financial QA but weaker on qualitative
questions \citep{mansouri2026mcp}; FinRank targets the retrieval-based
setting that remains necessary when answers must be grounded in the
filings themselves rather than a pre-structured data feed.

Across this body of work, evidence annotation identifies what is
relevant --- as gold passages or, in FinAgentBench's case, graded chunk
relevance --- but no benchmark ships, per question, a dedicated set of
human-selected, semantically confusable \emph{hard-negative passages}
against which discrimination can be scored directly against random
distractors. FinRank fills exactly this gap
(Table~\ref{tab:prior-work-comparison}). Each question carries gold
supporting passages \emph{and} human-selected distractors from comparable
filings, together with difficulty, reasoning-type, and evidence-scope
metadata and, for \QRCoveragePct\% of records, a gold sub-question decomposition,
so that retrieval, reranking, and hard-negative discrimination become
separately measurable. The nearest resource with curated hard negatives,
DocReRank \citep{wasserman2025docrerank}, generates hard-negative
\emph{queries} to train page-image rerankers, and is neither a QA
benchmark nor passage-level. Concurrent method-side work reaches the same
diagnosis from the systems direction: FinCARDS \citep{zhou2026fincards}
recasts financial evidence selection as constraint satisfaction over
entities, metrics, fiscal periods, and numeric spans, reranking
within-filing BM25 candidates on FinAgentBench
\citep{choi2025finagentbench} and reporting lexical retrieval to be
vulnerable to numeric drift, temporal misalignment, and boilerplate
repetition. FinCARDS contributes a method and releases code but no
evaluation data or negatives, and confines itself to intra-document
retrieval within a single filing, noting that its effectiveness across
multiple documents remains unevaluated. FinRank is complementary on both
counts: it supplies the annotated positives and curated distractors that
such constraint-aware rerankers are scored against, and its hard
negatives are drawn predominantly from \emph{other} filings, targeting
precisely the cross-document setting left open there.

\begin{table*}[t]
\centering
\small
\setlength{\tabcolsep}{5pt}
\adjustbox{max width=\linewidth}{%
\begin{tabular}{lllcccc}
\toprule
Benchmark & Source documents & \#\,Records & Retrieval & Hard neg. & Reranking & Rich metadata \\
\midrule
FinQA \citep{chen2021finqa}                & Earnings reports (S\&P~500)       & 8{,}281            & --      & -- & -- & -- \\
TAT-QA \citep{zhu2021tatqa}                & Report table/text snippets        & 16{,}552           & --      & -- & -- & -- \\
ConvFinQA \citep{chen2022convfinqa}        & Earnings reports (dialogue)       & 14{,}115           & --      & -- & -- & -- \\
FinanceBench \citep{islam2023financebench} & 10-K/10-Q/8-K/earnings            & 10{,}231$^\dagger$ & partial & -- & -- & sector \\
DocFinQA \citep{reddy2024docfinqa}         & Full 10-K documents               & 7{,}437            & yes     & -- & -- & -- \\
FinDER \citep{choi2025finder}              & 10-K (S\&P~500)                   & 5{,}703            & yes     & -- & yes$^\ddagger$ & topic \\
FinAgentBench \citep{choi2025finagentbench} & 10-K/10-Q/8-K, calls, DEF-14A     & 26{,}000$^{\S}$    & yes     & -- & yes$^{\S}$ & -- \\
\midrule
\textbf{FinRank (ours)}                    & 10-K/10-Q                         & \CorpusRecords{}   & \textbf{yes} & \textbf{yes} & \textbf{yes} & \textbf{full} \\
\bottomrule
\end{tabular}%
}
\caption{FinRank against representative financial QA and retrieval
benchmarks. ``Hard neg.''\ denotes per-question curated hard-negative
passages; ``Reranking'' a gold-vs.-hard-negative reranking task; ``Rich
metadata'' per-record difficulty, reasoning-type, and evidence-scope
labels. $^\dagger$150 questions are open-sourced.
$^\ddagger$FinDER evaluates LLM reranking of retrieved passages but
releases no curated hard-negative candidate set.
$^{\S}$FinAgentBench provides 26K graded-relevance annotations for
document- and chunk-ranking, without a per-question curated
hard-negative set. To our knowledge, FinRank is the first financial QA
benchmark to release per-question curated hard negatives and to score
discrimination against curated versus random distractors directly.}
\label{tab:prior-work-comparison}
\end{table*}

\subsection{Retrieval, Reranking, and Hard Negatives}

BM25 \citep{robertson2009bm25} remains a strong sparse baseline; dense
retrievers
\citep{karpukhin2020dpr,reimers2019sentencebert,izacard2022contriever}
and cross-encoder rerankers \citep{nogueira2019passage} form the standard
two-stage pipeline evaluated by general IR benchmarks such as BEIR
\citep{thakur2021beir} and MTEB \citep{muennighoff2023mteb}, with FinMTEB
\citep{tang2025finmteb} the financial counterpart. A long line of work
shows that \emph{hard} negatives (not random in-batch ones) are what
most improve dense retrievers
\citep{xiong2021ance,qu2021rocketqa,zhan2021hardneg}. FinRank carries this
insight to the evaluation side: rather than mining hard negatives to
train a retriever, it releases human-curated hard negatives as a fixed
benchmark asset and reports how much they degrade ranking relative to
random distractors (Section~\ref{sec:results}).

\subsection{Attribution and Faithfulness}

Retrieval-augmented generation \citep{lewis2020rag} can still hallucinate
or misattribute, motivating metrics for citation quality
\citep{gao2023alce}, reference-free RAG evaluation \citep{es2023ragas},
and atomic factual precision \citep{min2023factscore}. These score
\emph{generated} text; FinRank is complementary, measuring the retrieval
and reranking steps upstream through gold-versus-hard-negative passage
labels, so that evidence attribution can be assessed before generation is
introduced.

\section{Dataset Construction}
\label{sec:dataset-construction}

\subsection{Sources and Design Principles}
\label{sec:design-principles}

FinRank aggregates \CorpusRecords{} records, each grounded in a 10-K (annual) or
10-Q (quarterly) filing submitted to the SEC. The records were collected
by trained annotators, each responsible for roughly 250 records,
whose assignments together span the pharmaceutical, oil-and-gas, and
automotive sectors (Section~\ref{sec:annotation-pipeline}). The benchmark
is organized around four principles: examples are grounded in primary
disclosures rather than secondary summaries; every question is paired with
supporting passages for evidence-based evaluation; hard negatives are
included to support retrieval and reranking; and metadata exposes
variation across sector, topic, reasoning type, filing year, and document
type, so that performance can be analyzed beyond aggregate scores.

\subsection{Question-Generation Criteria}
\label{sec:methodology}

The annotators worked from a shared criteria document that fixes
what a well-formed FinRank record must contain; we summarize it here and
reproduce it in full in Appendix~\ref{app:methodology}. The criteria
define the unit of evidence (a \emph{passage}: a self-contained span of
filing text), separate \emph{qualitative} reasoning (interpretation and
judgment) from \emph{quantitative} reasoning (numerical computation,
itself split into metrics-generated, single-step, and compositional
calculations), and require, for medium- and hard-difficulty questions, a
\texttt{query\_rewrite} that decomposes the question into an ordered
sequence of atomic sub-questions.

Coverage is controlled along three axes with explicit targets. \emph{Topic:}
questions span eight filing-relevant domains (Company Overview,
Financials, Footnotes, Governance, Accounting, Legal, Risk, and
Shareholder Return), each mapped to a distinct set of 10-K/10-Q
disclosure items. \emph{Complexity:} a 30\%/40\%/30\% Easy/Medium/Hard
split, where Easy questions are direct factual lookups, Medium questions
require comparison or trend analysis across sections, and Hard questions
require multi-section synthesis. \emph{Evidence scope:} a 40\%/30\%/30\%
single-/two-/multi-passage split, so the benchmark is not dominated by
single-paragraph lookups. The criteria additionally instruct annotators
to balance qualitative and quantitative questions.
Section~\ref{sec:target-vs-realized} reports how closely the released
dataset meets each target.

Fixing these criteria \emph{ex ante} is what lets FinRank test more than
surface lexical matching: a system must locate the disclosure passages
that support the \emph{reasoning type} a question demands, not merely
passages that share its vocabulary, and, for multi-passage
questions, must integrate evidence spread across non-adjacent filing
sections.

\subsection{Record Schema}

The dataset is distributed as a single \texttt{.jsonl} file, one JSON
object per line. Each record is flat at the top level, with the question,
reference answer, optional \texttt{query\_rewrite} array, question- and
filing-level metadata, and two nested arrays: \texttt{passages}
(supporting passages, each with text and page number) and
\texttt{hard\_negatives} (each additionally tagged with its originating
\texttt{ticker}, \texttt{year}, and \texttt{doc\_type}). The full
field-by-field schema is given in Appendix~\ref{app:schema}.

\subsection{Supporting Passages and Hard Negatives}
\label{sec:hard-negatives}

The \texttt{passages} array contains the passages from the underlying
SEC filing that support the reference answer, each with a \texttt{text}
field and a \texttt{page\_number} field; some questions are answered by
a single passage, others require evidence drawn from multiple parts of
the same filing. Supporting passages thus play two roles: they are
ground-truth retrieval targets, and they constrain answer generation by
providing the evidence to which a faithful generator should be grounded.

Each entry in \texttt{hard\_negatives} additionally records the
originating filing's \texttt{ticker}, \texttt{year}, and
\texttt{doc\_type}, so distractors may come from another company,
another year, or a different filing type than the supporting evidence.
We treat hard negatives as a first-class benchmark asset rather than as
optional augmentation: the dataset is intended for explicit comparison
of retrievers and rerankers on their ability to suppress these
distractors.

\subsection{Data Collection}
\label{sec:annotation-pipeline}

FinRank was collected by five business students, each independently
responsible for a set of companies within one of the three sectors,
together spanning the
pharmaceutical, oil-and-gas, and automotive sectors. All annotators worked
from a shared question-generation criteria document
(Section~\ref{sec:methodology}; reproduced in
Appendix~\ref{app:methodology}) that fixes the topic taxonomy, the
qualitative and quantitative reasoning categories, the complexity-level
split, the passage-coverage split, and the query-rewrite rules. Every
record was authored \emph{manually}: students read the underlying 10-K
and 10-Q filings directly and wrote each question, its reference
answer, and, for medium- and hard-difficulty questions, the
\texttt{query\_rewrite} decomposition into sub-questions by hand. No
language model was used to generate questions, answers, or rewrites;
the templated \texttt{question\_id} strings (for example,
\texttt{JNJ\_\allowbreak CompanyOverview\_\allowbreak Hard\_\allowbreak MultiPassage\_\allowbreak Quantitative\_\allowbreak 03})
are the naming convention students followed from the criteria document,
not an artifact of automated generation.

Evidence was assembled by hand as well. For each question the
responsible student located the supporting passage or passages in the
source filing and transcribed them into the \texttt{passages} array
with their page references. The student then selected the record's
hard negatives manually, drawing plausible-but-incorrect passages from
comparable filings, most often a competitor in the same sector, and
otherwise the same company in a different reporting period or filing
type. Because hard negatives were chosen by a human reader searching
for the most confusable evidence rather than sampled by a retrieval
heuristic, they concentrate in the ``same industry, different company''
bucket that dominates Figure~\ref{fig:hn-taxonomy}. This distribution is
a deliberate annotation choice, and it is precisely the configuration
that most often confuses lexical and embedding-based retrievers
(Section~\ref{sec:results}).

During collection the authors performed a sampled review of each
student's records and returned corrective feedback, so that systematic
misreadings of the criteria could be caught before release. This review
was applied to a sample rather than exhaustively, and we did not compute
a formal inter-annotator-agreement statistic; both points are recorded
as limitations in Section~\ref{sec:limitations}. After collection we
applied a single deterministic normalization pass
(Section~\ref{sec:official-labels}; Appendix~\ref{app:normalization})
that harmonizes label surface forms, recovers the small number of
placeholder values deterministically from the structured
\texttt{question\_id}, and repairs transposed hard-negative metadata,
preserving every raw value in a parallel \texttt{<field>\_raw} field.

\subsection{Retrieval Corpus Definitions}
\label{sec:retrieval-corpus}

To make retrieval and reranking results on FinRank reproducible, we
specify two evaluation regimes that are both reproducible from the
JSONL alone, with no additional corpus required.

\paragraph{Global pooled corpus $C$.}
$C$ is the union of every non-empty supporting-passage text and
hard-negative text appearing anywhere in FinRank, deduplicated on the
raw passage text. Each entry is tagged with the
\texttt{(ticker, year, doc\_type, page\_number)} metadata of the first
record in which it appeared. For the released split, $|C| = \CorpusEntries$
unique passages drawn from \CorpusRecords{} records, with on average
\CorpusGoldMean{} supporting passages and \CorpusHnMean{} hard negatives
per record. First-stage retrieval baselines (Section~\ref{sec:results}) rank
all of $C$ for each query and report Recall@$k$, MRR, and nDCG@$10$
against the per-record gold positives. We emphasize what this regime is:
because $C$ consists solely of annotator-selected positives and curated
distractors, retrieval over $C$ is a \emph{curated passage-ranking
stress test} with an elevated density of confusable passages, not
retrieval through the complete text of each filing. Scores on $C$ are
therefore not comparable to full-document retrieval settings, and
extending FinRank with exhaustive filing-level chunks is left to a
future release.

\paragraph{In-record candidate set $L_r$.}
For each record $r$, $L_r = G_r \cup \mathrm{HN}_r$ is the union of
$r$'s supporting passages and its hard negatives. Reranking and
hard-negative-discrimination evaluations are restricted to $L_r$ and
report MRR, nDCG@$5$, and pairwise accuracy on
$(\text{positive}, \text{hard negative})$ pairs. This isolates ranking
quality from first-stage recall and matches how hard negatives are
intended to be used (Section~\ref{sec:hard-negatives}).

Studies using FinRank should state which regime is being reported for
each metric, and any corpus extension beyond $C$ (for example, a larger
filing-derived pool) should be released alongside the results.

\subsection{Hard-Negative Construction Taxonomy}
\label{sec:hn-taxonomy}

\begin{figure}[t]
\centering
\includegraphics[width=\linewidth]{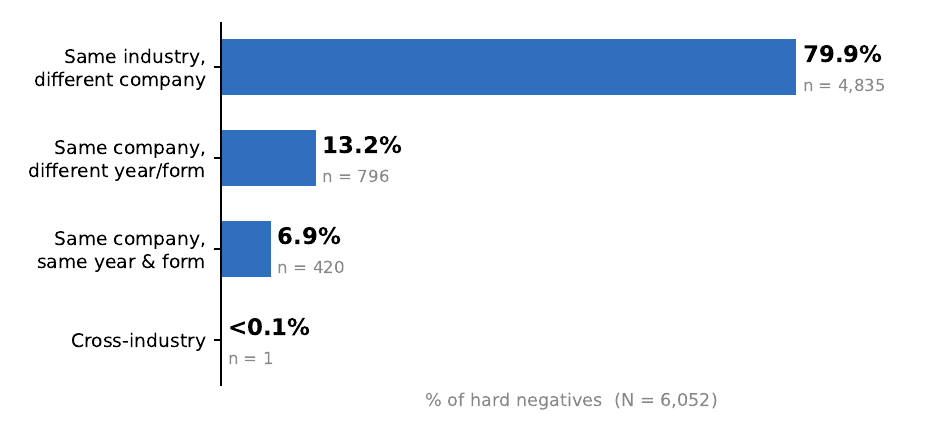}
\caption{Taxonomy of the \TaxTotal{} hard negatives, by the relationship
between each negative's originating filing and its source record. A hard
negative is most often a passage from a competitor's filing of comparable
type and period, the configuration retrievers are most likely to confuse
with true evidence.}
\label{fig:hn-taxonomy}
\end{figure}

Each hard negative carries the \texttt{(ticker, year, doc\_type)} of its
originating filing, so comparing it against the source record yields the
four-bucket taxonomy in Figure~\ref{fig:hn-taxonomy}. Because negatives
were selected by hand (Section~\ref{sec:annotation-pipeline}), 80\% fall
in the ``same industry, different company'' bucket: a competitor's
filing of comparable type and period, the case retrievers are most likely
to confuse with true evidence. A byte-equal audit identified 11 hard negatives that duplicated a
supporting passage of their own record; these, together with one
degenerate entry, are removed from the release. The remaining
\TaxContamHN{} hard negatives ($\sim$7.3\%) coincide with a supporting
passage of a \emph{different} record, where the passage may legitimately
be relevant to more than one question. We release \texttt{hn\_taxonomy.json} so that
users of hard-negative metrics can filter these overlaps.

\subsection{Official Label Set and Normalization}
\label{sec:official-labels}

To make evaluation reproducible without forcing every consumer to
re-derive the same mapping, the released \texttt{FinRank.jsonl}
carries the \emph{official benchmark labels}: a deterministic,
idempotent curation pass (Appendix~\ref{app:normalization})
title-cases \texttt{difficulty}, consolidates \texttt{industry} variants
into three sectors with the oil-and-gas sub-industry surfaced in a new
field, unifies \texttt{topic}, \texttt{passage\_type}, and
\texttt{reasoning\_type} typos, reconciles the \texttt{Ford}/\texttt{F}
ticker split, and recovers literal placeholders from the structured
\texttt{question\_id}. Every value the pass changed is preserved inline
in a parallel \texttt{<field>\_raw} field, every change is enumerated in
the released machine-readable repair log, and all distributional
statistics in this paper are reported on the canonical labels.

\section{Dataset Analysis}
\label{sec:dataset-analysis}

\begin{figure*}[t]
\centering
\includegraphics[width=.8\textwidth]{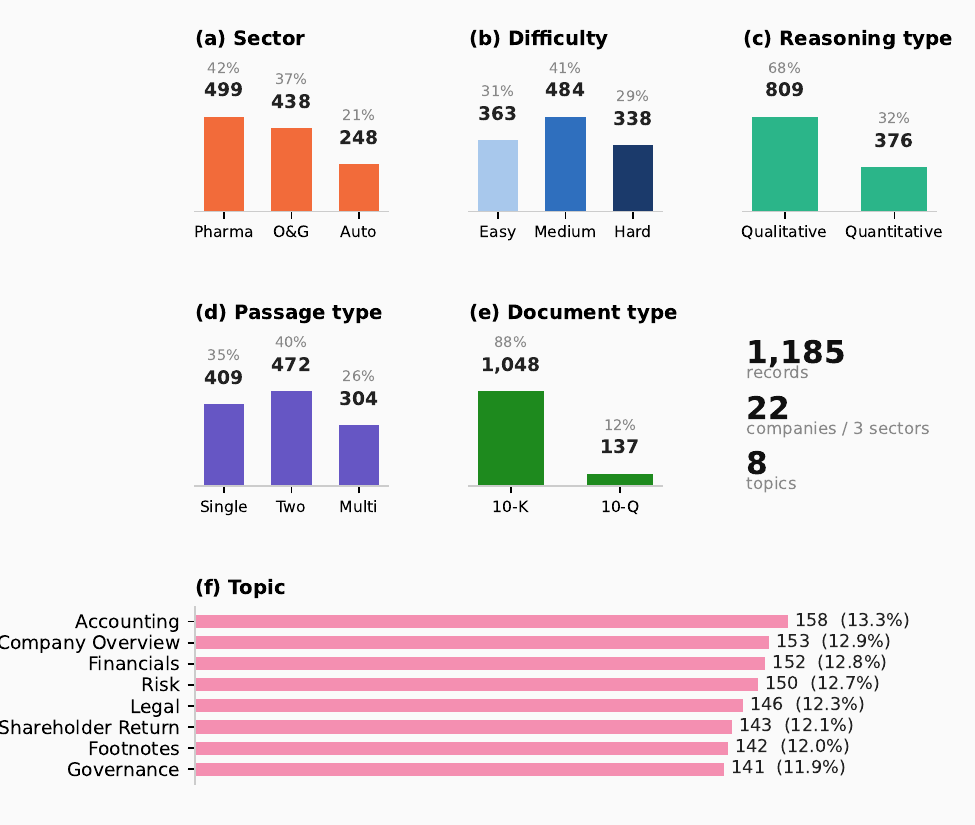}
\caption{Composition of FinRank across the \CorpusRecords{} records: (a) sector,
(b) difficulty, (c) reasoning type, (d) evidence scope (passage type),
(e) document type, and (f) topic. Counts are on the canonical normalized
labels. The dataset is deliberately near-uniform over topics
(141--158 records each) but skewed along sector, document type, and
reasoning type, skews that motivate stratified reporting.}
\label{fig:composition}
\end{figure*}

We summarize the composition of FinRank on the canonical normalized
labels (Section~\ref{sec:official-labels}), verified directly against the
released \texttt{FinRank.jsonl}. Figure~\ref{fig:composition} gives the
full picture along six axes; per-field distribution tables and the
raw-vs.-normalized counts are provided in Appendix~\ref{app:normalization}.

The \CorpusRecords{} records come from 22 companies in three sectors:
pharmaceuticals (\NormIndustryPharmaceuticals{} records), oil and gas
(\NormIndustryOilgas{}), and automotive (\NormIndustryAutomotive{}),
with per-company counts ranging from 9 (XOM) to 142 (JNJ).
Difficulty follows the intended 30/40/30 Easy/Medium/Hard split
(Figure~\ref{fig:composition}b), and topics are close to uniform
(Figure~\ref{fig:composition}f). Three axes are markedly skewed: 88\% of
records come from 10-K filings and only 12\% from 10-Qs; 75\% of filings
are from 2025 and 25\% from 2024; and qualitative questions outnumber
quantitative ones roughly two to one (68\% vs.\ 32\%). Evidence scope is
spread across single- (34\%), two- (40\%), and multi-passage (26\%)
records, so roughly two thirds of the benchmark require integrating
evidence from more than one part of a filing.

These skews mean that aggregate metrics on FinRank primarily reflect
performance on the dominant strata (10-K, 2025, qualitative, pharma and
oil-and-gas). We therefore recommend that retrieval and answer-generation
metrics be reported per stratum, by document type, filing year,
reasoning type, and sector, in addition to in aggregate, and we note
that the 2024 and 10-Q partitions are small and produce noisier
per-stratum estimates.

\subsection{Data Quality and Normalization}
\label{sec:data-quality-issues}

The data as collected contains surface-level label inconsistencies
(casing and spelling variants and a small number of literal template
placeholders) and, in $\sim$0.3\% of hard-negative entries, transposed
metadata fields. A deterministic, idempotent curation pass
(Section~\ref{sec:official-labels}; Appendix~\ref{app:normalization})
resolves these into the canonical labels of the released
\texttt{FinRank.jsonl}; original values are preserved in parallel
\texttt{<field>\_raw} fields, every change is enumerated in the released
repair log; duplicate passages within a record, degenerate hard
negatives, and annotation-tool artifacts embedded in passage text are
removed; and sixty-five records are excluded from the release entirely
(reliance on a non-filing source, placeholder or synthetic content, or
duplicate question--answer pairs), each enumerated in the repair log. Records
whose stored labels disagree with their structured \texttt{question\_id}
are flagged in the repair log for the audit discussed in
Section~\ref{sec:limitations}.

\subsection{Methodology-to-Dataset Validation}
\label{sec:target-vs-realized}

Because the question-generation criteria of Section~\ref{sec:methodology}
specify explicit target distributions over difficulty, passage coverage,
and reasoning type, we can measure realized deviations between the stated
targets and the released dataset (Table~\ref{tab:target-vs-realized}, on
canonical normalized labels).

\begin{table*}[t]
\centering
\small
\adjustbox{max width=\linewidth}{%
\begin{tabular}{p{3.6cm}p{2.4cm}p{2.4cm}p{2.4cm}}
\toprule
Dimension & Target & Realized & Deviation \\
\midrule
\multicolumn{4}{l}{\emph{Complexity}} \\
\quad Easy   & 30\% & 30.6\% & $+0.6$\,pp \\
\quad Medium & 40\% & 40.8\% & $+0.8$\,pp \\
\quad Hard   & 30\% & 28.5\% & $-1.5$\,pp \\
\midrule
\multicolumn{4}{l}{\emph{Passage coverage}} \\
\quad Single-passage & 40\% & 34.5\% & $-5.5$\,pp \\
\quad Two-passage    & 30\% & 39.8\% & $+9.8$\,pp \\
\quad Multi-passage  & 30\% & 25.7\% & $-4.3$\,pp \\
\midrule
\multicolumn{4}{l}{\emph{Reasoning type} (methodology calls for balance)} \\
\quad Qualitative  & balanced & 68.3\% & skewed (\,$\sim 2{:}1$\,) \\
\quad Quantitative & balanced & 31.7\% & skewed \\
\bottomrule
\end{tabular}%
}
\caption{Target distributions mandated by the FinRank
question-generation criteria (Section~\ref{sec:methodology}) against
realized values in the canonical normalized release. The 30/40/30
complexity target is met to within about one percentage point per
bucket; the 40/30/30 passage-coverage target deviates most, with
two-passage records over-represented; and reasoning type shows the
largest gap, with qualitative records outnumbering quantitative ones
roughly two to one. Consumers should treat these realized values as the
authoritative distribution and stratify results accordingly.}
\label{tab:target-vs-realized}
\end{table*}

The complexity split is met almost exactly. Passage coverage deviates
more: two-passage records are over-represented by about ten percentage
points, drawn from the single- and multi-passage buckets. Reasoning
type is the largest departure from the criteria, at roughly two
qualitative records per quantitative one. Appendix~\ref{app:cross-tab}
reports cross-tabulations showing that this qualitative skew is present
at every difficulty level.

\section{Tasks}
\label{sec:task-definition}

FinRank supports five evaluation tasks that mirror the stages of an
analyst workflow over SEC filings: locating candidate evidence
(retrieval), distinguishing it from plausible distractors (reranking),
integrating evidence across passages (multi-passage reasoning),
producing an answer faithful to that evidence (answer generation), and
mapping a colloquial query to a retrievable form (query rewriting).
Retrieval tasks operate on the global pooled corpus $C$ defined in
Section~\ref{sec:retrieval-corpus}; reranking and hard-negative
discrimination operate on the per-record candidate set $L_r$ from the
same section. Table~\ref{tab:tasks-overview} summarizes the inputs,
outputs, suggested metrics, and possible baselines for each task.

\begin{table*}[t]
\centering
\small
\adjustbox{max width=\linewidth}{%
\begin{tabular}{p{2.4cm}p{3.0cm}p{2.6cm}p{3.6cm}p{3.4cm}}
\toprule
Task & Inputs & Output & Metrics & Possible baselines \\
\midrule
Answer generation & Question; passages (gold or retrieved) & Natural-language answer & EM, token F1, ROUGE-L, BERTScore, faithfulness vs.\ provided passages & Closed-book LLM; open-book LLM; RAG \\
Passage retrieval & Question; passage corpus & Ranked candidate passages & Recall@$k$, MRR, nDCG@$k$, Hit Rate & TF-IDF, BM25 \citep{robertson2009bm25}; DPR \citep{karpukhin2020dpr}; SBERT \citep{reimers2019sentencebert}; hybrid \\
Reranking with hard negatives & Question; in-record positives and hard negatives & Ranking over the union & MRR, nDCG@$k$, pairwise accuracy on (positive, hard negative) pairs & Cross-encoder rerankers \citep{nogueira2019passage} \\
Multi-passage reasoning & Question; multiple supporting passages & Answer integrating all relevant passages & EM, F1, ROUGE-L, BERTScore, evidence coverage & Long-context LLMs; multi-passage RAG; question decomposition \\
Query rewriting & Original question & One or more reformulations & Semantic preservation (downstream answer agreement); retrieval improvement (Recall@$k$, MRR) & LLM-based and template-based rewriters \\
\bottomrule
\end{tabular}%
}
\caption{Tasks supported by FinRank, with inputs, outputs, suggested evaluation metrics, and possible baselines.}
\label{tab:tasks-overview}
\end{table*}

\paragraph{Citation and attribution evaluation.} Reranking with hard
negatives extends naturally to RAG-style citation evaluation: a system
that produces an answer together with one or more cited passages can
be scored on whether the cited passages overlap with the gold
supporting passages and whether they avoid the hard-negative set.
This is particularly relevant in financial settings where
auditability of generated answers is a practical concern.

\paragraph{Multi-passage records and the \texttt{query\_rewrite} field.}
Records with \texttt{passage\_type} \texttt{Two-Passage} or
\texttt{Multi-Passage} (Section~\ref{sec:dataset-analysis}) provide the
substrate for the multi-passage reasoning task. The
\texttt{query\_rewrite} array, where present, supplies reference
reformulations that can be used as targets for the query-rewriting
task or as paraphrase inputs for retrieval-robustness analyses.

\section{Experimental Protocol and Baselines}
\label{sec:experiments}

We report executed retrieval, reranking, and hard-negative discrimination
baselines using deterministic, publicly available models; generative
answer-quality evaluation is left to future work. All reported metrics
are produced by the released evaluation harness (\texttt{baselines/})
applied to the released \texttt{FinRank.jsonl} and are exactly
reproducible from these artifacts. All baselines are
evaluated on the full \CorpusRecords{}-record dataset using the canonical
normalized labels, and we fix \texttt{seed=42} for the only
non-deterministic component (random-negative sampling).
Appendix~\ref{app:reproducibility} lists dependencies and the reporting
checklist we recommend for future studies.

\subsection{Baselines}
\label{sec:baselines}

We run seven systems: \textbf{TF-IDF} (scikit-learn
\texttt{TfidfVectorizer}, sublinear TF, cosine scoring); \textbf{BM25}
(\texttt{rank-bm25} \texttt{BM25Okapi}, default parameters,
\citealp{robertson2009bm25}); a \textbf{dense retriever}
(\texttt{all-mpnet-base-v2}, \citealp{reimers2019sentencebert}, cosine
over L2-normalized embeddings); and a \textbf{cross-encoder reranker}
(\texttt{ms-marco-MiniLM-L-6-v2}, \citealp{nogueira2019passage}), applied
both over the top-20 dense candidates from the global pool $C$ and over
the in-record set $L_r$; a stronger general-purpose embedder
(\texttt{bge-large-en-v1.5}); a finance-adapted embedder
(\texttt{FinLang/finance-embeddings-investopedia}); and a 7B
instruction-tuned embedder (\texttt{e5-mistral-7b-instruct}, encoded
per its reference implementation), all under a uniform 512-token
truncation. We additionally report a
\textbf{metadata-filtered BM25} baseline: BM25 restricted to the corpus
entries whose \texttt{(ticker, year, doc\_type)} match the query
record's source filing, quantifying the effect of metadata
pre-filtering before semantic ranking. Baselines left to future work
include an SEC-filing-adapted retriever, a closed-book LLM, an
open-book LLM with gold passages, and a full RAG pipeline
\citep{lewis2020rag}.

\subsection{Splits and Ablations}
\label{sec:ablation-protocol}

The baselines here are computed over the entire dataset as reference
numbers. We additionally release five generalization splits as
record-ID assignments (\texttt{baselines/splits.json}): random
(\SplitRandomTrain/\SplitRandomDev/\SplitRandomTest{}
train/dev/test), held-out-ticker
(\SplitTickerTrain/\SplitTickerDev/\SplitTickerTest), leave-one-sector-out
folds, year (test $=$ 2024, \SplitYearTest{} records), and document
type (test $=$ 10-Q, \SplitDoctypeTest{} records). We recommend that future
studies report under at least one, noting that the year and 10-Q test
partitions are small and yield noisier estimates. We stratify results along document type,
reasoning type, passage type, and difficulty (Section~\ref{sec:results}),
and report two design-validating contrasts: a \emph{query-rewrite}
ablation (question alone vs.\ question concatenated with its
\texttt{query\_rewrite} sub-questions) and a \emph{hard-vs.-random}
negative contrast that tests whether the released hard negatives are in
fact harder than arbitrary distractors.

\section{Results}
\label{sec:results}

The retrieval, reranking, and hard-negative discrimination tables
below report numbers actually produced by the baselines of
Section~\ref{sec:baselines} on the entire \CorpusRecords-record
dataset. Numbers are emitted by
\texttt{baselines/make\_macros.py} from the JSON output of the run
scripts, and any rerun automatically refreshes them. Answer generation
and faithfulness are left to future work
(Section~\ref{sec:experiments}).

\begin{figure*}[t]
\centering
\includegraphics[width=.8\textwidth]{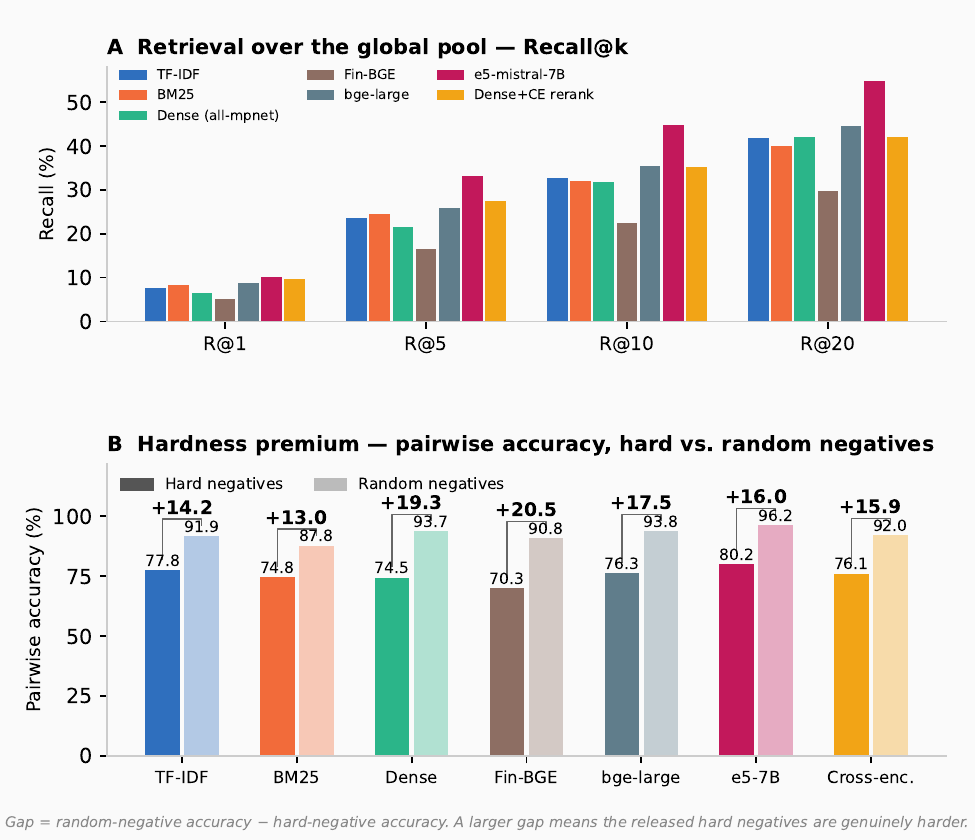}
\caption{Executed baselines on FinRank. \textbf{(A)} First-stage
retrieval over the global pool: Recall@$k$ for the evaluated systems.
Even the strongest, a 7B instruction-tuned embedder, reaches only
\EFiveMistralRecallTen\% Recall@10, and sub-billion-parameter dense
encoders gain little over BM25. \textbf{(B)} The hardness premium:
pairwise ranking accuracy of
each model against its curated hard negatives versus random negatives.
The 13.0--20.5 point gap is the empirical evidence that the released hard
negatives are genuinely harder than arbitrary distractors.}
\label{fig:results}
\end{figure*}

\subsection{Retrieval over the Global Pool}

Table~\ref{tab:retrieval-overall} and Figure~\ref{fig:results}(A) report
first-stage retrieval over the global pooled corpus $C$
(Section~\ref{sec:retrieval-corpus}; $|C| = \CorpusEntries$). Each
score is the mean across queries against the per-record gold positives.

\begin{table*}[t]
\centering
\small
\adjustbox{max width=\linewidth}{%
\begin{tabular}{lcccccc}
\toprule
Model & Recall@1 & Recall@5 & Recall@10 & Recall@20 & MRR & nDCG@10 \\
\midrule
TF-IDF                       & \TfidfRecallOne   & \TfidfRecallFive   & \TfidfRecallTen   & \TfidfRecallTwenty   & \TfidfMRR   & \TfidfNDCG \\
BM25                         & \BMRecallOne      & \BMRecallFive      & \BMRecallTen      & \BMRecallTwenty      & \BMMRR      & \BMNDCG \\
Dense (all-mpnet-base-v2)    & \DenseRecallOne   & \DenseRecallFive   & \DenseRecallTen   & \DenseRecallTwenty   & \DenseMRR   & \DenseNDCG \\
\quad + CE rerank of mpnet top-20 & \CeDenseRecallOne & \CeDenseRecallFive & \CeDenseRecallTen & \CeDenseRecallTwenty & \CeDenseMRR & \CeDenseNDCG \\
Dense (finance-adapted BGE)  & \FinBgeRecallOne  & \FinBgeRecallFive  & \FinBgeRecallTen  & \FinBgeRecallTwenty  & \FinBgeMRR  & \FinBgeNDCG \\
Dense (bge-large-en-v1.5)    & \BgeLargeRecallOne & \BgeLargeRecallFive & \BgeLargeRecallTen & \BgeLargeRecallTwenty & \BgeLargeMRR & \BgeLargeNDCG \\
Dense (e5-mistral-7b-instruct) & \EFiveMistralRecallOne & \EFiveMistralRecallFive & \EFiveMistralRecallTen & \EFiveMistralRecallTwenty & \EFiveMistralMRR & \EFiveMistralNDCG \\
\bottomrule
\end{tabular}%
}
\caption{Retrieval performance on the global pooled corpus $C$
($|C|=\CorpusEntries$, $\CorpusRecords$ queries). All values are
percentages averaged over queries.}
\label{tab:retrieval-overall}
\end{table*}

\paragraph{Metadata-filtered BM25.}
Because most hard negatives originate from a different filing, a system
that knows the target filing could pre-filter candidates on metadata
before ranking. We quantify this directly: restricting $C$ to
entries whose \texttt{(ticker, year, doc\_type)} match the query
record's filing eliminates \MetaFilterHnElimPct\% of that record's hard
negatives and shrinks the pool to $\approx$\MetaFilterMeanCands{}
candidates on average, lifting BM25 from \BMRecallTen{} to
\MetaFilterBmRecallTen{} Recall@10 (MRR \BMMRR{} $\rightarrow$
\MetaFilterBmMRR{}). Because pooled entries carry first-occurrence
metadata (Section~\ref{sec:retrieval-corpus}), the filter excludes at
least one gold passage for \MetaFilterGoldLossSome{} records and every
gold passage for \MetaFilterGoldLossAll{}; occurrence-level source
attribution is left to a future release. Two readings
follow. First, metadata filtering is genuinely powerful, and deployed
filing assistants should use it; results on FinRank's unfiltered pool
characterize the complementary regime in which filing attribution is
unavailable or unreliable (noisy metadata, cross-filing questions,
multi-company corpora). Second, even under this filter, nearly
half of the gold evidence is missed at $k=10$: within-filing semantic
discrimination, the setting probed by the same-filing and same-company
hard negatives, is not eliminated by metadata filtering.

\subsection{Reranking over the In-Record Candidate Set}

Table~\ref{tab:reranking-results} reports reranking on the per-record
candidate set $L_r = G_r \cup \mathrm{HN}_r$
(Section~\ref{sec:retrieval-corpus}), isolating ranking quality from
first-stage recall.

\begin{table}[t]
\centering
\small
\adjustbox{max width=\linewidth}{%
\begin{tabular}{lcc}
\toprule
Model & MRR & nDCG@5 \\
\midrule
TF-IDF                          & \TfidfInRecMRR  & \TfidfInRecNDCGFive \\
BM25                            & \BMInRecMRR     & \BMInRecNDCGFive \\
Dense (all-mpnet-base-v2)       & \DenseInRecMRR  & \DenseInRecNDCGFive \\
Dense (finance-adapted BGE)     & \FinBgeInRecMRR & \FinBgeInRecNDCGFive \\
Dense (bge-large-en-v1.5)       & \BgeLargeInRecMRR & \BgeLargeInRecNDCGFive \\
Dense (e5-mistral-7b-instruct)  & \EFiveMistralInRecMRR & \EFiveMistralInRecNDCGFive \\
Cross-encoder over $L_r$        & \CeInrecInRecMRR & \CeInrecInRecNDCGFive \\
\bottomrule
\end{tabular}%
}
\caption{Reranking performance on the in-record candidate set $L_r$.}
\label{tab:reranking-results}
\end{table}

\subsection{Hard-Negative Discrimination}

Figure~\ref{fig:results}(B) reports pairwise accuracy across all
$(\text{positive}, \text{hard negative})$ pairs per record under the
in-record ranking, contrasted with the same metric against random
negatives drawn uniformly from the pooled corpus $C$ excluding the
record's own positives (Section~\ref{sec:ablation-protocol}, seed 42,
one random negative per curated hard negative; tied scores count for
the positive, and records are averaged with equal weight). Every model ranks a positive above a
\emph{random} negative 88--96\% of the time but above a curated
\emph{hard} negative only 70--80\% of the time, a 13.0--20.5 point drop.
This gap is the empirical hardness premium of the released hard
negatives, and it is the single result that most directly justifies
shipping them as a first-class benchmark asset.

\subsection{Stratified Performance and Ablations}
\label{sec:stratified}

Table~\ref{tab:results-stratified} reports retrieval and reranking
metrics stratified along four axes (document type, reasoning type,
passage type, difficulty). Per-stratum partition sizes are emitted
alongside the metrics in the underlying \texttt{ablations.json}.
Selected cells are reproduced below; the full stratified breakdown is
available in \texttt{baselines/results/ablations.json}.

\begin{table*}[t]
\centering
\small
\adjustbox{max width=\linewidth}{%
\begin{tabular}{llccc}
\toprule
Axis & Stratum & BM25 R@10 & Dense R@10 & CE rerank R@10 \\
\midrule
\multirow{2}{*}{Document type}
  & 10-K          & \AblDoctypeTenKBmTwofiveRecallTen     & \AblDoctypeTenKDenseRecallTen     & \AblDoctypeTenKCeDenseTopKRecallTen \\
  & 10-Q          & \AblDoctypeTenQBmTwofiveRecallTen     & \AblDoctypeTenQDenseRecallTen     & \AblDoctypeTenQCeDenseTopKRecallTen \\
\midrule
\multirow{2}{*}{Reasoning}
  & Qualitative   & \AblReasoningtypeQualitativeBmTwofiveRecallTen   & \AblReasoningtypeQualitativeDenseRecallTen   & \AblReasoningtypeQualitativeCeDenseTopKRecallTen \\
  & Quantitative  & \AblReasoningtypeQuantitativeBmTwofiveRecallTen  & \AblReasoningtypeQuantitativeDenseRecallTen  & \AblReasoningtypeQuantitativeCeDenseTopKRecallTen \\
\midrule
\multirow{3}{*}{Passage type}
  & Single-Passage & \AblPassagetypeSinglepassageBmTwofiveRecallTen & \AblPassagetypeSinglepassageDenseRecallTen & \AblPassagetypeSinglepassageCeDenseTopKRecallTen \\
  & Two-Passage    & \AblPassagetypeTwopassageBmTwofiveRecallTen    & \AblPassagetypeTwopassageDenseRecallTen    & \AblPassagetypeTwopassageCeDenseTopKRecallTen \\
  & Multi-Passage  & \AblPassagetypeMultipassageBmTwofiveRecallTen  & \AblPassagetypeMultipassageDenseRecallTen  & \AblPassagetypeMultipassageCeDenseTopKRecallTen \\
\midrule
\multirow{3}{*}{Difficulty}
  & Easy           & \AblDifficultyEasyBmTwofiveRecallTen   & \AblDifficultyEasyDenseRecallTen   & \AblDifficultyEasyCeDenseTopKRecallTen \\
  & Medium         & \AblDifficultyMediumBmTwofiveRecallTen & \AblDifficultyMediumDenseRecallTen & \AblDifficultyMediumCeDenseTopKRecallTen \\
  & Hard           & \AblDifficultyHardBmTwofiveRecallTen   & \AblDifficultyHardDenseRecallTen   & \AblDifficultyHardCeDenseTopKRecallTen \\
\bottomrule
\end{tabular}%
}
\caption{Recall@10 stratified by document type, reasoning type, passage
type, and difficulty. Per-stratum sample sizes are available in
\texttt{ablations.json}.}
\label{tab:results-stratified}
\end{table*}

\paragraph{Query-rewrite ablation.}
Table~\ref{tab:queryrewrite-ablation} compares retrieval performance
on the subset of records with non-empty \texttt{query\_rewrite} lists
(\QRCount{} records; see \texttt{ablations.json}) under two conditions: the
raw question alone, and the raw question concatenated with all
listed rewrites.

\begin{table}[t]
\centering
\small
\adjustbox{max width=\linewidth}{%
\begin{tabular}{lcc}
\toprule
Model / condition & Recall@10 (raw) & Recall@10 (with rewrites) \\
\midrule
TF-IDF & \QRRawTfidfRecallTen   & \QRWithrewritesTfidfRecallTen \\
BM25   & \QRRawBmTwofiveRecallTen & \QRWithrewritesBmTwofiveRecallTen \\
Dense  & \QRRawDenseRecallTen   & \QRWithrewritesDenseRecallTen \\
\bottomrule
\end{tabular}%
}
\caption{Query-rewrite ablation on the subset of records with
non-empty \texttt{query\_rewrite} lists. Because rewrites are
annotator-authored decompositions written with the gold evidence in
view, the right column should be read as an oracle upper bound on
decomposition-assisted retrieval. The right column concatenates
the original question with all listed rewrites; the left uses the
question alone.}
\label{tab:queryrewrite-ablation}
\end{table}

\subsection{Answer Generation and Error Analysis}

FinRank is designed to support answer-generation evaluation
(closed-book, open-book with gold passages, and full RAG, scored with
EM, F1, ROUGE-L, BERTScore, and citation-faithfulness metrics
\citep{gao2023alce,es2023ragas}) and a qualitative error analysis that
separates retrieval failures (no supporting passage in the top-$k$),
reranking failures (a hard negative preferred to a supporting passage),
generation failures on correctly retrieved passages, and attribution
failures. Because these require a generative model and a human-validated
or LLM-judged correctness protocol, we leave them to future work and
report only the executed retrieval and reranking results above; the
evaluation protocol is specified in Section~\ref{sec:experiments} so
that such studies remain comparable.

\section{Discussion}
\label{sec:discussion}

\paragraph{What the baselines reveal.} Four observations stand out from
Section~\ref{sec:results}. First, ranking within a curated pool of
confusable filing passages is hard: the strongest evaluated system, a
7B instruction-tuned embedder, reaches 44.8\% Recall@10 over the pooled
corpus $C$, leaving more than half of the gold evidence outside the
top ten. Because
$C$ contains only annotated positives and curated distractors rather
than the full text of each filing, these numbers measure discrimination
among confusable disclosure passages, not end-to-end retrieval over
complete documents; full-document retrieval remains untested here. Second, encoder capacity dominates domain labels on this heavily
templated text: sub-billion-parameter encoders earn little (mpnet ties
BM25; bge-large gains 3.5 Recall@10 points), a finance-adapted embedder
tuned on consumer-finance text trails BM25 by 9.7 points, while the 7B
instruction-tuned e5-mistral gains 12.7 points over BM25 --- scale and
instruction tuning help where wrong-register domain adaptation hurts,
and SEC-filing-specific adaptation remains open. Third, and most
directly tied to the benchmark's design, the released hard negatives cost
every model 13.0--20.5 points of pairwise accuracy relative to random
distractors; this hardness premium is the single number that justifies
shipping curated hard negatives as a first-class asset. Fourth, the
stratified breakdown exposes where aggregate scores hide weakness:
10-Q, multi-passage, and quantitative records score lowest, which is
precisely why we recommend per-stratum reporting. The 10-Q gap should
be read descriptively rather than causally: all 10-Q records come from
the oil and gas sector and predominantly from 2025
(Appendix~\ref{app:cross-tab}), so document type is confounded with
sector, filing year, and annotator, and disentangling these effects
requires the controlled splits of Section~\ref{sec:ablation-protocol}.

\paragraph{Why evidence grounding matters in finance.} In audit,
compliance, and investor-relations settings the value of an answer is
tied to its provenance: a reviewer needs to know not just \emph{what} a
system claims but \emph{where} the claim comes from, so that the
underlying disclosure can be inspected. FinRank operationalizes this by
making the identification of supporting passages a first-class evaluation
target and by pairing it with hard negatives that stress the boundary
between genuinely supporting evidence and plausible distractors. Because
those distractors are most often a competitor's comparable filing
(Section~\ref{sec:hn-taxonomy}), the benchmark rewards systems that
disambiguate between lexically and topically similar disclosures, the
failure mode most likely to matter in practice. Multi-passage records add
a complementary axis, since financial conclusions frequently combine an
MD\&A narrative, a footnote, and a tabular disclosure.

\paragraph{Scope.} FinRank is intended for evaluating evidence retrieval,
reranking, and (in future work) answer generation for financial-analyst
assistants, compliance QA tools, and SEC-filing search interfaces. It is
not designed or validated for automated investment decisions, valuation,
or trading signals, and any deployment in high-stakes settings should
retain human oversight.

\section{Limitations}
\label{sec:limitations}

\paragraph{Scale and skew.} FinRank contains \CorpusRecords{} records over 22
companies and three sectors, enough for ranking and answer-quality
comparisons, but small relative to open-domain QA benchmarks, and skewed
across sector, filing year (75\% from 2025), document type (87\% 10-K),
and reasoning type (roughly $2{:}1$ qualitative). Aggregate metrics
therefore reflect the dominant strata, per-stratum subsets can be noisy,
and performance should not be assumed to transfer to companies, sectors,
jurisdictions, or periods outside the release.

\paragraph{Single-annotator review.} Records were authored manually by the
student annotators and reviewed on a sampled basis by the authors
during collection (Section~\ref{sec:annotation-pipeline}); we did not
re-verify every record or compute a formal inter-annotator-agreement
statistic, because each record was authored by a single student rather
than double-annotated. Correctness on FinRank is thus single-annotator
with sampled review rather than consensus-adjudicated, and a stratified
double-annotation study is the most important outstanding item for a
future release.

\paragraph{Label and hard-negative artifacts.} Raw fields contain casing
and spelling variants and a few placeholder values, which our
normalization pipeline (Section~\ref{sec:official-labels}) resolves;
consumers who skip normalization may see spurious \texttt{Unknown}
buckets. Separately, \TaxContamHN{} hard negatives ($\sim$7.3\%) are
byte-equal to a supporting passage of a \emph{different} record, where
the passage may legitimately be relevant to more than one question; the
released \texttt{hn\_taxonomy.json} lets users filter these overlaps.

\paragraph{Modality and generation.} Tables and figures are represented as
text, so evaluations requiring true multimodal reasoning over image-based
tables are out of scope, and reference-based metrics may penalize answers
that are correct but phrased differently from the reference. Finally, this
paper reports only retrieval and reranking baselines; generative
answer-quality results are left to future work, so no claim about
generator performance on FinRank follows from it.

\section{Ethical Considerations}
\label{sec:ethics}

FinRank is derived from filings that companies have submitted to the SEC
and made public; it contains no non-public personal information. As is
inherent to SEC filings, passages may name executives, directors, and
officers in their public corporate roles, including disclosures such as
executive compensation that issuers are legally required to publish.
FinRank is released under CC~BY-NC~4.0 for non-commercial research use
with attribution; commercial licensing is available from the authors
(Appendix~\ref{app:dataset-card}). Because the dataset is skewed toward a
few sectors, filing years, and document types
(Section~\ref{sec:dataset-analysis}), models tuned or evaluated on it may
reflect the language of the dominant strata, and cross-sector
generalization should be tested explicitly rather than inferred from
aggregate scores.

Strong performance on FinRank does not establish suitability for
investment decisions, valuation, or trading, and FinRank-aligned systems
should complement rather than replace qualified human analysts. Even
accurate retrieval and generation systems can mis-rank passages, omit
relevant evidence, or produce plausible but incorrect answers, so
generated answers should not be presented as investment advice and should
carry inline citations to their supporting passages, supporting the
audit and compliance review that financial disclosure requires.

\section{Conclusion}
\label{sec:conclusion}

We introduced FinRank, an evidence-grounded benchmark for financial
question answering and retrieval over SEC 10-K and 10-Q filings. Across
\CorpusRecords{} manually authored records spanning 22 companies, three sectors,
and eight topics for filings from 2024--2025, each question is paired with
supporting passages from the underlying filing and with a curated set of
hard negatives from comparable filings, making FinRank, to our knowledge, the
first financial QA benchmark to release per-question curated hard
negatives and to measure discrimination against curated versus random
distractors directly. Among the evaluated baselines, even a 7B instruction-tuned embedder
reaches only 44.8\% Recall@10 on the curated evidence pool,
sub-billion-parameter embeddings beat BM25 by at most 3.5 points on
this templated text, and the released hard negatives cost every model
13.0--20.5 points of pairwise accuracy over random distractors. Future work
includes broadening coverage to more sectors, companies, and filing years;
a stratified double-annotation study with reported inter-annotator
agreement; executing the generation and RAG protocols with LLM-based
systems; and extending the schema and hard-negative construction to
multimodal (tabular and graphical) evidence.

\clearpage
\printbibliography

\appendix
\onecolumn
\section{Question Generation and Annotation Criteria}
\label{app:methodology}

This appendix consolidates the full question-generation methodology
that governs FinRank, complementing the summary in
Section~\ref{sec:methodology}. The methodology fixes the criteria
against which the released JSONL was constructed and against which a
third party can audit or extend the benchmark.

\subsection{Purpose}

The methodology defines criteria for generating questions about
financial filings so that the resulting question set provides
systematic coverage across financial topics, reasoning types, and
complexity levels. The intent is a balanced and thorough evaluation
resource, not an arbitrary collection of natural-language prompts.

\subsection{Core Definitions}

\paragraph{Passage.} A discrete section of text from a financial
filing that can provide standalone context for answering a question.
Passages are the unit of evidence and the unit of retrieval.

\paragraph{Financial filing.} An SEC document, specifically a Form
10-K (annual report) or Form 10-Q (quarterly report), containing a
company's financial and operational disclosures.

\paragraph{Query rewrite.} A decomposition of a complex question into
a sequence of sub-questions that guide analytical reasoning and
expose intermediate retrieval and reasoning steps.

\subsection{Reasoning Types}

\paragraph{Qualitative reasoning.} Interpretation, explanation, or
judgment questions that do not require arithmetic calculations
(e.g.\ business strategy analysis, revenue-recognition judgments,
litigation impact discussions).

\paragraph{Quantitative reasoning.} Any question requiring numerical
computation, comparison, or mathematical analysis. Quantitative
questions are further partitioned into three subtypes:
\begin{enumerate}[leftmargin=*,itemsep=2pt]
    \item \textbf{Metrics-generated}: questions derived automatically
    from standardized financial ratios and templated calculations
    (e.g.\ current-ratio queries).
    \item \textbf{Single-step arithmetic}: a single division,
    multiplication, addition, or subtraction.
    \item \textbf{Compositional calculations}: multi-step arithmetic
    requiring two or more sequential operations.
\end{enumerate}

\subsection{Financial Topic Categories}

The methodology partitions filing content into eight financial topic
categories. For each topic the methodology supplies both qualitative
and quantitative reasoning targets (Table~\ref{tab:topic-reasoning-matrix}).

\begin{table}[h]
\centering
\small
\adjustbox{max width=\linewidth}{%
\begin{tabular}{p{2.4cm}p{5.4cm}p{5.4cm}}
\toprule
Topic & Qualitative reasoning examples & Quantitative reasoning examples \\
\midrule
Company Overview   & Business nature, strategy evaluation, M\&A rationale & YoY revenue growth, headcount changes, EBITDA margin \\
Financials         & Earnings quality, operating margin consistency       & EPS changes, debt-to-equity, free cash flow per share \\
Footnotes          & Accounting policy implications, lease classifications & Depreciation schedules, stock-compensation expense \\
Governance         & Board composition, audit committee scope             & Director independence \%, executive pay ratios \\
Accounting         & Revenue-recognition judgments, impairment triggers   & Amortization changes, deferred-tax calculations \\
Legal              & Contingent liability discussions, litigation impacts  & Legal provision amounts, legal cost trends \\
Risk               & Risk sensitivity analysis, risk management evaluation & Value-at-Risk, interest coverage ratios \\
Shareholder Return & Dividend policy rationale, buyback strategy           & Total shareholder returns, dividend payout ratios \\
\bottomrule
\end{tabular}%
}
\caption{Topic--reasoning matrix from the FinRank question-generation
methodology. The matrix is prescriptive: it specifies the
$(\text{topic}, \text{reasoning type})$ cells that the dataset is
expected to populate.}
\label{tab:topic-reasoning-matrix}
\end{table}

\subsection{Complexity-Level Definitions}

Questions are distributed across three complexity levels with target
shares $30\%$ Easy, $40\%$ Medium, $30\%$ Hard.

\paragraph{Easy (direct / factual).} Simple factual retrieval from a
filing section, straightforward numerical queries, basic definitional
questions; minimal interpretation. Examples: ``What was the company's
total revenue in 2023?''; ``What is the company's primary business
segment?''; ``When does the company's fiscal year end?''

\paragraph{Medium (analytical).} Comparisons between periods or
segments, basic trend analysis, percentage calculations, and
cross-referencing between related filing sections. Medium-level
questions ship with a multi-sub-question \texttt{query\_rewrite}.
Example original question: ``How did the company's profitability
change between 2022 and 2023, and what drove this change?''
Rewrite (abridged): gross/operating/net margins for both periods,
followed by an attribution sub-question for the year-over-year change.

\paragraph{Hard (complex synthesis).} Multi-step reasoning across
multiple sections, complex synthesis, risk assessment using multiple
data points, or strategic analysis combining various filing
elements. Hard-level questions ship with three or more sub-questions
that sequentially reduce the synthesis problem to factual queries.
Example original question: ``How do the company's liquidity risks
relate to their debt covenant requirements and upcoming maturities?''
Rewrite (abridged): current, quick, and cash ratios; cash balances;
specific covenant requirements; current covenant margins; 12- and
24-month maturities; final synthesis sub-question.

\subsection{Query Rewrite Rules}

Query rewrites serve four purposes: they break complexity into
manageable components, ensure completeness of the reasoning chain,
make the analytical process transparent, and allow evaluation of
intermediate retrieval steps rather than only end-to-end answer
correctness. The application rule is:
\begin{itemize}[leftmargin=*,itemsep=2pt]
    \item \textbf{Easy} questions: generally do not require rewrites.
    \item \textbf{Medium} questions: require multiple atomic sub-questions.
    \item \textbf{Hard} questions: require three or more sub-questions
    that sequentially decompose complex reasoning into factual queries.
\end{itemize}

\subsection{Passage Coverage Requirements}

Questions are also distributed by the number of passages required to
answer them, with target shares $40\%$ single-passage ($1$ passage),
$30\%$ two-passage (exactly $2$ passages), and $30\%$ multi-passage
($3$ or more passages). The intent is to force the benchmark to
include multi-passage synthesis, which more closely matches realistic
financial analysis.

\subsection{Implementation Guidelines}

The methodology binds the above criteria into five implementation
guidelines for any party generating FinRank-style questions:
\begin{enumerate}[leftmargin=*,itemsep=2pt]
    \item \textbf{Balance across topics}: every topic category is
    represented.
    \item \textbf{Reasoning-type mix}: balance between qualitative and
    quantitative questions.
    \item \textbf{Complexity distribution}: adhere to the
    $30\%$--$40\%$--$30\%$ split.
    \item \textbf{Passage requirements}: follow the
    $40\%$--$30\%$--$30\%$ split exactly.
    \item \textbf{Sub-question creation}: generate the required
    rewrites for medium and hard questions.
\end{enumerate}
Realized deviations from these guidelines are reported in
Table~\ref{tab:target-vs-realized}.

\subsection{Related Resources}

The methodology cites two prior financial-QA resources as conceptual
context for FinRank-style question generation. FinanceBench
\citep{islam2023financebench} provides a comprehensive open-book QA
test suite of $10{,}231$ questions about publicly traded companies,
and the FinDER dataset of \citet{choi2025finder} contributes
expert-generated query--evidence--answer triplets with realistic
abbreviations and domain-specific language for RAG evaluation in
finance. FinRank differs from both in foregrounding the retrieval and
reranking sub-tasks with explicit hard negatives drawn from
comparable filings (Section~\ref{sec:related-work}).

\subsection{What the Methodology Does and Does Not Guarantee}
\label{app:methodology-guarantees}

The methodology provides structured \emph{generation criteria}: it
fixes what topics, reasoning types, complexity levels, passage
counts, and rewrites a FinRank question must satisfy, and is
therefore the source of truth for reproducibility and benchmark
validity arguments. It does \emph{not} by itself certify the
correctness of any individual question, answer, supporting passage,
or hard negative; it does not specify human validation, expert
adjudication, or inter-annotator agreement statistics. We treat the
methodology as evidence that the dataset's generation procedure was
deliberate and as the basis for the deviation reporting in
Table~\ref{tab:target-vs-realized}, while recording label-correctness
audits as outstanding items in Section~\ref{sec:limitations}.

\section{Record Schema}
\label{app:schema}

Table~\ref{tab:schema} gives the top-level schema of a FinRank record.

\begin{table}[h]
\centering
\small
\adjustbox{max width=\linewidth}{%
\begin{tabular}{p{2.6cm}p{2.0cm}p{8.5cm}}
\toprule
Field & Type & Description \\
\midrule
\texttt{id} & string & UUID identifier of the record. \\
\texttt{question\_id} & string & Original question identifier inherited from the source folder. \\
\texttt{ticker} & string & Stock ticker of the filing company. \\
\texttt{industry} & string & Industry classification associated with the filing company. \\
\texttt{question} & string & Natural-language question. \\
\texttt{answer} & string & Reference answer text. \\
\texttt{query\_rewrite} & array & Optional alternative formulations of the question. \\
\texttt{topic} & string & Topical category of the question (for example, Risk, Accounting, Governance). \\
\texttt{reasoning\_type} & string & Qualitative or Quantitative reasoning required to answer. \\
\texttt{difficulty} & string & Easy, Medium, or Hard, with some casing variation in the raw data. \\
\texttt{num\_passages} & string/int & Number of supporting passages required to answer. \\
\texttt{passage\_type} & string & Single-, One-, Two-, or Multi-Passage indicator, with some spelling variation in the raw data. \\
\texttt{year} & string/int & Filing year of the underlying document. \\
\texttt{doc\_type} & string & SEC document type (10-K or 10-Q). \\
\texttt{passages} & array & Supporting passages with text and page reference. \\
\texttt{hard\_negatives} & array & Hard negative passages with text, page reference, ticker, year, and document type. \\
\texttt{folder\_name} & string & Source folder of the record. \\
\bottomrule
\end{tabular}%
}
\caption{Top-level schema of a FinRank record.}
\label{tab:schema}
\end{table}

The nested \texttt{passages} and \texttt{hard\_negatives} objects,
abbreviated above, are detailed in
Tables~\ref{tab:passages-schema} and~\ref{tab:hard-negatives-schema}.

\begin{table}[h]
\centering
\small
\adjustbox{max width=\linewidth}{%
\begin{tabular}{p{2.6cm}p{2.0cm}p{8.5cm}}
\toprule
Field & Type & Description \\
\midrule
\texttt{text} & string & Passage content drawn from the underlying SEC filing. \\
\texttt{page\_number} & string/int & Page reference within the source document. \\
\texttt{passage\_id} & string & Stable identifier \texttt{<record id>:p<index>} (normalized release). \\
\texttt{text\_sha1} & string & SHA-1 of the whitespace-trimmed text (normalized release). \\
\bottomrule
\end{tabular}%
}
\caption{Schema of an entry in the \texttt{passages} array.}
\label{tab:passages-schema}
\end{table}

\begin{table}[h]
\centering
\small
\adjustbox{max width=\linewidth}{%
\begin{tabular}{p{2.6cm}p{2.0cm}p{8.5cm}}
\toprule
Field & Type & Description \\
\midrule
\texttt{text} & string & Hard-negative passage content. \\
\texttt{page\_number} & string/int & Page reference within the source document. \\
\texttt{ticker} & string & Ticker of the company whose filing the hard negative is drawn from. \\
\texttt{year} & string/int & Filing year of the source document. \\
\texttt{doc\_type} & string & Source document type (for example, 10-K or 10-Q). \\
\texttt{passage\_id} & string & Stable identifier \texttt{<record id>:hn<index>} (normalized release). \\
\texttt{text\_sha1} & string & SHA-1 of the whitespace-trimmed text (normalized release). \\
\bottomrule
\end{tabular}%
}
\caption{Schema of an entry in the \texttt{hard\_negatives} array.}
\label{tab:hard-negatives-schema}
\end{table}

\section{Canonical Label Normalization Mapping}
\label{app:normalization}

Table~\ref{tab:normalization-mapping} enumerates the canonical label
normalizations applied in curating the released
\texttt{FinRank.jsonl} (Section~\ref{sec:official-labels}). Changed
values are preserved inline in parallel \texttt{<field>\_raw} fields,
and every individual change is enumerated in the released
\texttt{repair\_log.json}.

\begin{table}[h]
\centering
\small
\adjustbox{max width=\linewidth}{%
\begin{tabular}{lll}
\toprule
Field & Raw value(s) & Canonical value \\
\midrule
\texttt{reasoning\_type} & \texttt{Qaltitative}, \texttt{Qualtitative}, \texttt{Qalitative} & \texttt{Qualitative} \\
\texttt{reasoning\_type} & \texttt{"Quantitative~"} (trailing space) & \texttt{Quantitative} \\
\texttt{reasoning\_type} & \texttt{"string"} & Recovered from \texttt{question\_id} \\
\texttt{passage\_type} & \texttt{One-Passage}, \texttt{One\_Passage}, \texttt{OnePassage} & \texttt{Single-Passage} (merged) \\
\texttt{passage\_type} & \texttt{Tow-Passage}, \texttt{Two-Page} & \texttt{Two-Passage} \\
\texttt{passage\_type} & \texttt{Mutli-Passage}, \texttt{Multi\_Passage}, \texttt{Multi-passage} & \texttt{Multi-Passage} \\
\texttt{passage\_type} & \texttt{"string"} & Recovered from \texttt{question\_id} \\
\texttt{topic} & \texttt{Foodnotes} & \texttt{Footnotes} \\
\texttt{topic} & \texttt{CompanyOverview} & \texttt{Company Overview} \\
\texttt{topic} & \texttt{ShareholderReturn} & \texttt{Shareholder Return} \\
\texttt{topic} & \texttt{Financial} & \texttt{Financials} \\
\texttt{industry} & \texttt{Pharmaceutical}, \texttt{Phamaceuticals} & \texttt{Pharmaceuticals} \\
\texttt{industry} & \texttt{Oil and Gas} & \texttt{Oil \& Gas} \\
\texttt{industry} & \texttt{Oil \& Gas \{Midstream,E\&P,...\}} & \texttt{Oil \& Gas} (sub-industry surfaced) \\
\texttt{difficulty} & lowercase / leading-space variants & Title-case Easy / Medium / Hard \\
\texttt{ticker} & \texttt{Ford} & \texttt{F} \\
\texttt{ticker} & \texttt{"string"} & Recovered from \texttt{question\_id} \\
\texttt{difficulty} & \texttt{"string"} & Recovered from \texttt{question\_id} \\
\texttt{doc\_type} & multi-source compounds (2 records) & \texttt{Unknown} \\
\texttt{year} & \texttt{"integer"} & Majority year of same-ticker, same-\texttt{doc\_type} records (flagged as inferred) \\
\bottomrule
\end{tabular}%
}
\caption{Canonical label normalization mapping applied in curating the
release. Changed values are preserved in parallel
\texttt{<field>\_raw} fields.}
\label{tab:normalization-mapping}
\end{table}

Beyond label mapping, the pipeline performs three record-level repairs,
each preserved in a machine-readable repair log: (i) hard-negative
metadata whose fields were transposed at entry time is restored by rule
(a numeric \texttt{ticker} alongside an alphabetic \texttt{page\_number}
is swapped back, 10 cases; a \texttt{ticker}/\texttt{doc\_type}
rotation is reversed, 2 cases; one placeholder hard-negative year is
inferred by ticker--document majority); (ii) the 11 hard negatives whose
trimmed text duplicated a gold passage of their own record are removed,
as are empty and within-record duplicate passages, and page references
that are not page numbers are nulled;
and (iii) every stored label is cross-checked against the structured
\texttt{question\_id}, with the 56 disagreements shipped in the repair
log as candidates for manual audit rather than silently overwritten.
A final exclusion pass removes sixty-five records from the release:
fifty-two whose question, answer, rewrite, or evidence involves a
non-filing source (an academic journal article one annotator used as an
analytical frame, detected both by download-watermark markers and by
the article's distinctive regression variables); three whose hard
negatives are synthetic relevance rationales rather than filing
passages; three with placeholder or empty question or answer text; four
exact duplicates of another record's question--answer pair; and three
records left with fewer than four unique hard negatives. Each exclusion is recorded in
\texttt{repair\_log.json} with its rule; the release-blocking checks are
implemented in \texttt{baselines/validate\_release.py}.

\section{Example Record}
\label{app:example-record}

A sanitized and abbreviated example of a single FinRank record (long
passages and the full hard-negative array are truncated). The example
illustrates several data-quality observations from
Section~\ref{sec:data-quality-issues}: a \texttt{reasoning\_type} typo
(\texttt{Qaltitative}), a lower-case \texttt{difficulty}
(\texttt{easy}), and inconsistent \texttt{doc\_type} casing between
the supporting passages and the hard negatives.

\begin{verbatim}
{
  "id": "97546a79-...-82c119a9c38c",
  "question_id": "JNJ_CompanyOverview_Easy_OnePassage_Qualitative_01",
  "ticker": "JNJ",
  "industry": "Pharmaceuticals",
  "question": "Which reportable business segments does Johnson & Johnson
               operate in as of FY2024?",
  "answer": "As of FY2024, Johnson & Johnson operates in two main
             segments: Innovative Medicine and MedTech ...",
  "query_rewrite": [],
  "topic": "Company Overview",
  "reasoning_type": "Qaltitative",
  "difficulty": "easy",
  "num_passages": "1",
  "passage_type": "One-Passage",
  "year": "2025",
  "doc_type": "10-K",
  "passages": [
    {
      "text": "### Description of the company and business segments ...",
      "page_number": "22"
    }
  ],
  "hard_negatives": [
    {
      "text": "#### Note 19: Segment Information ...",
      "page_number": "106",
      "ticker": "LLY",
      "year": "2025",
      "doc_type": "10-k"
    }
  ],
  "folder_name": "annotator_01"
}
\end{verbatim}

\section{Reasoning-Type Examples}
\label{app:reasoning-examples}

To make the conceptual range of the questions concrete, we reproduce one
record per representative reasoning mode below, verbatim from
\texttt{FinRank.jsonl} (excerpts truncated for space). Each shows the
question, an excerpt of the reference answer, an excerpt of one
supporting passage, and an excerpt of one hard negative. The hard
negatives illustrate the dominant ``same industry, different company''
bucket of Section~\ref{sec:hn-taxonomy}.


\begin{tcolorbox}[enhanced,breakable,colback=black!2,colframe=black!30,boxrule=0.4pt,arc=1.5pt,left=5pt,right=5pt,top=4pt,bottom=4pt,fontupper=\footnotesize]
\textbf{Accounting interpretation}\hfill{\scriptsize\itshape JNJ, 2025 10-K \textperiodcentered\ Accounting \textperiodcentered\ Hard \textperiodcentered\ Qualitative}\par\vspace{2pt}
\textbf{Q.}~How do Johnson \& Johnson’s tangible and intangible assets contribute differently to the Company’s ability to generate revenue and sustain competitiveness, and what does this imply about future business risks and strategic priorities?\par\vspace{2pt}
\textbf{Reference answer.}~Johnson \& Johnson’s tangible assets, such as manufacturing facilities and equipment, are used efficiently, as seen in the higher PPE turnover ratio. This shows that the Company effectively leverages its physical asset base to support large-scale production and distribution. However, the balance sheet is far more heavily weighted toward in\,\ldots\par\vspace{2pt}
\textbf{Supporting passage}~(p.~44).~Johnson \& Johnson and subsidiaries consolidated balance sheets At December 29, 2024 and December 31, 2023 (Dollars in Millions Except Share and Per Share Amounts) (Note 1) 2024 2023 Assets Current assets Cash and cash equivalents\,\ldots\par\vspace{2pt}
\textbf{Hard negative}~(PFE, 2025 10-K).~Consolidated Balance Sheets Pfizer Inc. and Subsidiary Companies (MILLIONS, EXCEPT PER SHARE DATA) Assets As of December 31, 2024 2023 Cash and cash equivalents \$ 1,043 \$ 2,853 Short-term investments\,\ldots
\end{tcolorbox}
\vspace{3pt}

\begin{tcolorbox}[enhanced,breakable,colback=black!2,colframe=black!30,boxrule=0.4pt,arc=1.5pt,left=5pt,right=5pt,top=4pt,bottom=4pt,fontupper=\footnotesize]
\textbf{Risk-disclosure comparison}\hfill{\scriptsize\itshape JNJ, 2025 10-K \textperiodcentered\ Risk \textperiodcentered\ Hard \textperiodcentered\ Qualitative}\par\vspace{2pt}
\textbf{Q.}~What types of costs does Johnson \& Johnson report in its footnotes that illustrate risks from sudden market shifts, and how did the COVID-19 vaccine manufacturing exit generate such costs?\par\vspace{2pt}
\textbf{Reference answer.}~In its footnotes, Johnson \& Johnson discloses a variety of costs that arise when markets or business conditions change, such as litigation charges, restructuring costs, impairment of acquired assets, acquisition and integration expenses, divestiture-related losses, and regulatory compliance charges. These items show how rapidly evolving m\,\ldots\par\vspace{2pt}
\textbf{Supporting passage}~(p.~88-89).~(3) Innovative Medicine segment income before tax includes: • Acquired in-process research \& development expense of \$ 1.25billion to secure the global rights to the NM26 bispecific antibody (Yellow Jersey acquisition) • Monetizati\,\ldots\par\vspace{2pt}
\textbf{Hard negative}~(LLY, 2025 10-K).~Numbers may not add due to rounding. (1) Jardiance revenue includes Glyxambi, Synjardy, and Trijardy XR. (2) Humalog revenue includes insulin lispro. (3) Basaglar revenue includes Rezvoglar. (4) Olumi\,\ldots
\end{tcolorbox}
\vspace{3pt}

\begin{tcolorbox}[enhanced,breakable,colback=black!2,colframe=black!30,boxrule=0.4pt,arc=1.5pt,left=5pt,right=5pt,top=4pt,bottom=4pt,fontupper=\footnotesize]
\textbf{Quantitative extraction}\hfill{\scriptsize\itshape JNJ, 2025 10-K \textperiodcentered\ Financials \textperiodcentered\ Hard \textperiodcentered\ Quantitative}\par\vspace{2pt}
\textbf{Q.}~What was Johnson \& Johnson’s free cash flow to the firm (FCFF) in 2024, based on the reported financials?\par\vspace{2pt}
\textbf{Reference answer.}~Free cash flow to the firm is calculated by starting from operating profit after tax and adjusting for non-cash charges, capital expenditures, and working capital movements. The company reported EBIT of 20,804 million and with an effective tax rate of 15.7\% this results in a NOPAT of 17,536 million. Depreciation and amortization added bac\,\ldots\par\vspace{2pt}
\textbf{Supporting passage}~(p.~45).~Johnson \& Johnson and subsidiaries consolidated statements of earnings (Dollars and Shares in Millions Except Per Share Amounts) (Note 1) 2024 2023 2022 Sales to customers \$88,821 85,159 79,990 Cost of products sold 27,471 26,553\,\ldots\par\vspace{2pt}
\textbf{Hard negative}~(LLY, 2025 10-K).~ELI LILLY AND COMPANY AND SUBSIDIARIES <div align='center'>(Dollars in millions, except per-share data, and shares in thousands)</div> Year Ended December 31, 2024 2023 2022 Revenue (Note 2) \$ 45,042.\,\ldots
\end{tcolorbox}
\vspace{3pt}

\begin{tcolorbox}[enhanced,breakable,colback=black!2,colframe=black!30,boxrule=0.4pt,arc=1.5pt,left=5pt,right=5pt,top=4pt,bottom=4pt,fontupper=\footnotesize]
\textbf{Business-segment identification}\hfill{\scriptsize\itshape JNJ, 2025 10-K \textperiodcentered\ Company Overview \textperiodcentered\ Hard \textperiodcentered\ Qualitative}\par\vspace{2pt}
\textbf{Q.}~How has the company restructured its business areas in recent years, and what strategic rationale can be derived from its current focus on two segments?\par\vspace{2pt}
\textbf{Reference answer.}~Johnson \& Johnson has restructured its business in recent years by fully separating its Consumer Health segment through a three-step process: the Kenvue IPO in May 2023, an August 2023 exchange offer reducing ownership to 9.5\%, and a debt-for-equity exchange in Q2 2024 that eliminated the remaining stake. While transition manufacturing an\,\ldots\par\vspace{2pt}
\textbf{Supporting passage}~(p.~50).~On May 8, 2023, Kenvue, completed an initial public offering (the IPO) resulting in the issuance of 198,734,444 shares of its common stock, par value \$0.01 per share (the “Kenvue Common Stock”), at an initial public offering of \$2\,\ldots\par\vspace{2pt}
\textbf{Hard negative}~(LLY, 2025 10-K).~Divestitures Olanzapine Portfolio (including Zyprexa) In July 2023, we sold the rights for the olanzapine portfolio, including Zyprexa, to Cheplapharm Arzneimittel GmbH (Cheplapharm), a European compa\,\ldots
\end{tcolorbox}
\vspace{3pt}

\begin{tcolorbox}[enhanced,breakable,colback=black!2,colframe=black!30,boxrule=0.4pt,arc=1.5pt,left=5pt,right=5pt,top=4pt,bottom=4pt,fontupper=\footnotesize]
\textbf{Legal / regulatory interpretation}\hfill{\scriptsize\itshape ABBV, 2025 10-K \textperiodcentered\ Legal \textperiodcentered\ Hard \textperiodcentered\ Qualitative}\par\vspace{2pt}
\textbf{Q.}~How do AbbVie’s inherited Allergan litigations, combined with the financial aftermath of divested assets and integration costs, illustrate the legal and strategic risks that follow from large-scale acquisitions?\par\vspace{2pt}
\textbf{Reference answer.}~AbbVie’s acquisition of Allergan shows how legal exposures can extend well beyond the transaction date and shape the company’s financial risk profile. The company inherited hundreds of lawsuits, including opioid litigation and breast implant cases, where plaintiffs seek compensatory and punitive damages, medical monitoring, and other reme\,\ldots\par\vspace{2pt}
\textbf{Supporting passage}~(p.~96-97).~Government Proceedings Lawsuits are pending against Allergan and several other manufacturers generally alleging that they improperly promoted and sold prescription opioid products. Approximately 435lawsuits are pending against All\,\ldots\par\vspace{2pt}
\textbf{Hard negative}~(JNJ, 2025 10-K).~The Company is subject to significant legal proceedings that can result in significant expenses, fines and reputational damage. In the ordinary course of business, Johnson \& Johnson and its subsidiari\,\ldots
\end{tcolorbox}
\vspace{3pt}

\begin{tcolorbox}[enhanced,breakable,colback=black!2,colframe=black!30,boxrule=0.4pt,arc=1.5pt,left=5pt,right=5pt,top=4pt,bottom=4pt,fontupper=\footnotesize]
\textbf{Shareholder-return analysis}\hfill{\scriptsize\itshape LLY, 2025 10-K \textperiodcentered\ Shareholder Return \textperiodcentered\ Hard \textperiodcentered\ Quantitative}\par\vspace{2pt}
\textbf{Q.}~How closely does Eli Lilly’s stock performance align with its revenue and EPS growth trajectory, and what explains any divergence?\par\vspace{2pt}
\textbf{Reference answer.}~Eli Lilly’s stock performance has outpaced both revenue and EPS growth in recent years. Revenue increased by 20\% in 2023 and by 32\% in 2024, while diluted EPS declined by 16\% in 2023 due to higher internal and acquired R\&D expenses before rebounding with 102\% growth in 2024. By contrast, the stock delivered annualized gains of more than 4\,\ldots\par\vspace{2pt}
\textbf{Supporting passage}~(p.~40).~The following graph compares the return on Lilly stock with that of the Standard \& Poor's (S\&P) 500 Stock Index and our peer group for the years 2020 through 2024. The graph assumes that, on the last business day of 2019, a person\,\ldots\par\vspace{2pt}
\textbf{Hard negative}~(JNJ, 2025 10-K).~Johnson \& Johnson and subsidiaries consolidated statements of earnings (Dollars and Shares in Millions Except Per Share Amounts) (Note 1) 2024 2023 2022 Sales to customers \$88,821 85,159 79,990 Cost o\,\ldots
\end{tcolorbox}
\vspace{3pt}

\section{Example Prompt Templates}
\label{app:prompts}

We provide example prompt templates for three evaluation modes; benchmark
consumers should report the exact prompts used (including system
prompts and any few-shot exemplars) alongside their results.

\subsection{LLM-Based QA Prompt}

\begin{verbatim}
You are a financial analyst answering a question about a company's
SEC filing. Use only the information in the provided passages. If
the passages do not contain enough information to answer the
question, respond that the answer is not available in the
provided passages.

Question: {question}

Passages:
{passages}

Answer:
\end{verbatim}

\subsection{RAG-with-Citations Prompt}

\begin{verbatim}
You are a financial analyst answering a question about a company's
SEC filing. Use only the information in the provided passages. For
every claim in your answer, cite the passage that supports it
using its number, in square brackets, e.g. [1], [2]. Do not
include claims that are not supported by the provided passages.

Question: {question}

Passages:
[1] {passage_1}
[2] {passage_2}
...

Answer with citations:
\end{verbatim}

\subsection{Hard-Negative Discrimination Prompt}

\begin{verbatim}
You are evaluating which passages from a candidate set support
the answer to a financial question about an SEC filing. Read the
question and the candidate passages and assign each passage one
of two labels:

  - SUPPORTING: the passage contains evidence that directly
    supports an answer to the question.
  - NON-SUPPORTING: the passage does not directly support an
    answer to the question, even if it is topically related.

Return your decisions as a list of (passage_id, label) pairs.

Question: {question}

Candidate passages:
[1] {passage_1}
[2] {passage_2}
...

Decisions:
\end{verbatim}

\section{Reproducibility Notes}
\label{app:reproducibility}

In addition to the checklist in Section~\ref{sec:experiments}, studies
should release the exact split files (record \texttt{id}s per
partition) rather than describing splits in prose only, report whether
and how the normalization of Section~\ref{sec:official-labels}
was applied (with raw and normalized counts for each affected field),
report the verbatim prompts used by any LLM-based component, describe
how the retrieval corpus was assembled (in-record passages and hard
negatives only, or a broader corpus from the underlying filings), and
document model versions, decoding settings, and hardware to support
replication.

\paragraph{Auditability of the question set.}
Future users should be able to regenerate or audit the dataset using
the methodology document (Appendix~\ref{app:methodology}), the
question-generation criteria (Section~\ref{sec:methodology}), the
released \texttt{FinRank.jsonl}, the per-passage identifiers,
text hashes, and inline \texttt{<field>\_raw} values it carries, the
\texttt{query\_rewrite} arrays, the machine-readable repair log of the
normalization pipeline, and the evaluation scripts in
\texttt{baselines/}. Pairing the methodology with the realized
distribution (Table~\ref{tab:target-vs-realized}) is what allows a
third party to detect divergence from the documented criteria
without having to rederive them from scratch; it makes the benchmark
more transparent and reduces ambiguity in question design, while not
substituting for the question-by-question correctness audit listed
in Section~\ref{sec:limitations}.

\section{Cross-Tabulation Tables}
\label{app:cross-tab}

The following cross-tabulations of the FinRank metadata are useful
for diagnosing where future model performance is likely to be most
variable. All counts are computed on the canonical normalized labels
(Section~\ref{sec:official-labels}) by
\texttt{baselines/make\_crosstabs.py}; per-row and per-column
totals match the marginal distributions reported in
Section~\ref{sec:dataset-analysis}.

\begin{table}[h]
\centering
\small
\adjustbox{max width=\linewidth}{%
\begin{tabular}{lrrrrrrrrr}
\toprule
Industry & Accounting & Company Overview & Financials & Footnotes & Governance & Legal & Risk & Shareholder Return & Row total \\
\midrule
Automotive & 35 & 33 & 32 & 23 & 27 & 30 & 36 & 32 & 248 \\
Oil \& Gas & 60 & 55 & 53 & 57 & 55 & 55 & 51 & 52 & 438 \\
Pharmaceuticals & 63 & 65 & 67 & 62 & 59 & 61 & 63 & 59 & 499 \\
\midrule
Column total & 158 & 153 & 152 & 142 & 141 & 146 & 150 & 143 & 1185 \\
\bottomrule
\end{tabular}%
}
\caption{Cross-tabulation of industry by topic on the canonical normalized labels.}
\label{tab:ct-industry-topic}
\end{table}

\begin{table}[h]
\centering
\small
\adjustbox{max width=\linewidth}{%
\begin{tabular}{lrrr}
\toprule
Industry & 10-K & 10-Q & Row total \\
\midrule
Automotive & 248 & -- & 248 \\
Oil \& Gas & 301 & 137 & 438 \\
Pharmaceuticals & 499 & -- & 499 \\
\midrule
Column total & 1048 & 137 & 1185 \\
\bottomrule
\end{tabular}%
}
\caption{Cross-tabulation of industry by document type on the canonical normalized labels.}
\label{tab:ct-industry-doctype}
\end{table}

\begin{table}[h]
\centering
\small
\adjustbox{max width=\linewidth}{%
\begin{tabular}{lrrr}
\toprule
Filing year & 10-K & 10-Q & Row total \\
\midrule
2024 & 283 & 13 & 296 \\
2025 & 765 & 124 & 889 \\
\midrule
Column total & 1048 & 137 & 1185 \\
\bottomrule
\end{tabular}%
}
\caption{Cross-tabulation of filing year by document type on the canonical normalized labels. The Unknown entry is a record whose raw \texttt{doc\_type} names multiple source documents and cannot be attributed to a single filing.}
\label{tab:ct-year-doctype}
\end{table}

\begin{table}[h]
\centering
\small
\adjustbox{max width=\linewidth}{%
\begin{tabular}{lrrr}
\toprule
Passage type & Qualitative & Quantitative & Row total \\
\midrule
Single-Passage & 276 & 133 & 409 \\
Two-Passage & 322 & 150 & 472 \\
Multi-Passage & 211 & 93 & 304 \\
\midrule
Column total & 809 & 376 & 1185 \\
\bottomrule
\end{tabular}%
}
\caption{Cross-tabulation of passage type by reasoning type on the canonical normalized labels.}
\label{tab:ct-passagetype-reasoning}
\end{table}

\begin{table}[t]
\centering
\small
\adjustbox{max width=\linewidth}{%
\begin{tabular}{lrrr}
\toprule
Difficulty & Qualitative & Quantitative & Row total \\
\midrule
Easy & 229 & 134 & 363 \\
Medium & 348 & 136 & 484 \\
Hard & 232 & 106 & 338 \\
\midrule
Column total & 809 & 376 & 1185 \\
\bottomrule
\end{tabular}%
}
\caption{Cross-tabulation of difficulty by reasoning type on the canonical normalized labels. The qualitative-to-quantitative skew is present at every difficulty level.}
\label{tab:ct-difficulty-reasoning}
\end{table}

\paragraph{Observations.}
The (industry $\times$ topic) cross-tab
(Table~\ref{tab:ct-industry-topic}) shows that all three industries
populate every topic bucket, with no empty cell; Automotive cells are
smaller in absolute terms because Automotive has half as many records
as Pharmaceuticals or Oil~\&~Gas. The (year $\times$ doc\_type)
cross-tab (Table~\ref{tab:ct-year-doctype}) confirms that the 10-Q
partition is concentrated in a single filing year, so per-year and
per-doc\_type analyses are not independent. The (difficulty $\times$
reasoning\_type) and (passage\_type $\times$ reasoning\_type)
cross-tabs (Tables~\ref{tab:ct-difficulty-reasoning},
\ref{tab:ct-passagetype-reasoning}) expose where the
roughly $2{:}1$ qualitative-to-quantitative skew of
Section~\ref{sec:dataset-analysis} is most pronounced; consumers who
report difficulty- or passage-type-stratified metrics should also
report the corresponding qualitative/quantitative split rather than
assume it matches the dataset-level marginal.

\section{Dataset Card}
\label{app:dataset-card}

\paragraph{Identity.}
Name: \textbf{FinRank}. Distribution:
\url{https://github.com/datanxt/FinRank}. Distribution file:
\texttt{FinRank.jsonl}, carrying the canonical labels of
Section~\ref{sec:official-labels} with raw surface forms preserved
inline in \texttt{<field>\_raw} fields.

\paragraph{Contents.}
\CorpusRecords{} question--answer records derived from the 10-K and
10-Q filings of 22 publicly listed companies (three sectors, 8
topical labels, filing years 2024--2025). Each record contains a
question, a reference answer, optional query rewrites, structured
metadata, an array of supporting passages, and an array of hard
negative passages. The release additionally includes a deduplicated
global pooled retrieval corpus of \CorpusEntries{} unique passages
(see Section~\ref{sec:retrieval-corpus}).

\paragraph{Intended use.}
Benchmarking financial information-retrieval and question-answering
systems on SEC-filing prose, with explicit support for retrieval,
reranking, hard-negative discrimination, multi-passage reasoning,
query rewriting, and retrieval-augmented generation with citations.

\paragraph{Prohibited use.}
Automated investment decisions, automated trading signals,
real-time valuation, regulatory or compliance decisions, or any
high-stakes financial action without independent human oversight.
Outputs of any system evaluated on FinRank must not be presented as
financial advice. Reference answers reflect a particular reading of
forward-looking, ranged, or conditional disclosure language and are
not guaranteed to be correct in all financial respects.

\paragraph{Known issues.}
Label inconsistencies in the raw release (Section~\ref{sec:data-quality-issues});
distributional skews along industry, year, document type, and
reasoning type (Section~\ref{sec:dataset-analysis});
hard-negative contamination of \TaxContamHN{} HNs across
\TaxContamRecords{} records (Section~\ref{sec:hn-taxonomy}); no
independent ground-truth audit performed in this paper, see
Section~\ref{sec:limitations}.

\paragraph{Recommended evaluation practice.}
Use the released canonical labels for all
metric computation; report retrieval metrics on the global pool $C$
and reranking / hard-negative metrics on the in-record set $L_r$;
report stratified results in addition to aggregate results
(see Section~\ref{sec:stratified}); report the random seed and the
exact split used; and accompany generative answer-quality results
with faithfulness and citation-accuracy numbers in addition to
EM/F1/ROUGE-L/BERTScore.

\paragraph{License.}
FinRank is released under the Creative Commons
Attribution-NonCommercial 4.0 International license (CC~BY-NC~4.0).
Academic and educational use is permitted with attribution; commercial
licensing is available from the authors on request. The underlying
SEC filings are public records; FinRank redistributes only derived
question--answer records, supporting passages, and hard negatives.

\section{Baselines Reproduction}
\label{app:baselines-reproduction}

The retrieval, reranking, hard-negative, and ablation results in
Section~\ref{sec:results} are produced by the scripts in
\texttt{baselines/}. The complete pipeline is:

\begin{verbatim}
cd FinRank/baselines
pip install -r requirements.txt

python validate_release.py       --data ../FinRank.jsonl
python run_baselines.py          --data ../FinRank.jsonl --out results/
python run_extra_retrievers.py   --data ../FinRank.jsonl --out results/
python make_splits.py            --data ../FinRank.jsonl --out splits.json
python make_crosstabs.py         --data ../FinRank.jsonl --out ../
python make_macros.py            --data ../FinRank.jsonl \
    --results results/results.json \
    --repair-log ../repair_log.json --out _results_macros.tex
\end{verbatim}

Models and seeds:

\begin{itemize}
    \item Random seed: \texttt{42} (random-negative sampling).
    \item Dense retriever: \texttt{sentence-transformers/all-mpnet-base-v2}
    (\(\sim\)420\,MB).
    \item Cross-encoder: \texttt{cross-encoder/ms-marco-MiniLM-L-6-v2}
    (\(\sim\)80\,MB).
    \item Sparse: \texttt{rank-bm25} \texttt{BM25Okapi} with default
    parameters; scikit-learn \texttt{TfidfVectorizer} with sublinear
    TF and unigram features.
    \item Tokenization for sparse baselines: lowercase + punctuation
    stripping (regex \texttt{[\^{}a-z0-9]+}).
\end{itemize}

Runtime expectations: on a single A10G GPU the full harness
(\texttt{run\_baselines.py}) completes in under five minutes and the
additional dense retrievers (\texttt{run\_extra\_retrievers.py},
including the 7B \texttt{e5-mistral}) in roughly ten; on a CPU-only
laptop the mpnet/cross-encoder harness takes about an hour, and the 7B
model is impractical. Score matrices are cached to
\texttt{baselines/results/score\_cache.npz} so repeated runs do not
re-encode the corpus.

\end{document}